\documentclass{article} % For LaTeX2e
\usepackage{iclr2026_conference,times}
\usepackage{comment}
\usepackage{amssymb}
\usepackage{amsthm}
\usepackage[table]{xcolor}
\usepackage{mathtools}
\usepackage{xcolor}
\usepackage{graphicx}
\usepackage{tikz}
\usepackage{todonotes}
\usetikzlibrary{positioning,calc,fit,backgrounds}

\usepackage{amsmath,amsfonts,bm}

\def\eqref#1{equation~\ref{#1}}
\def\1{\bm{1}}

\DeclareMathAlphabet{\mathsfit}{\encodingdefault}{\sfdefault}{m}{sl}
\SetMathAlphabet{\mathsfit}{bold}{\encodingdefault}{\sfdefault}{bx}{n}

\usepackage{hyperref}
\usepackage{url}
\usepackage{booktabs}
\usepackage{multirow}

\title{Spectral-Target Physical Latent Structuring for JEPA-Style World Models}

\author{Penghao Zhu, 
Salvatore Penachio, 
Kaustav Mukherjee, 
\&  Aneesh Jonelagadda\thanks{Corresponding author. All authors contributed equally.} \\
Kaliber Labs\\
San Francisco, CA 94103, USA \\
\texttt{\{p.zhu,s.penachio,k.mukherjee,a.jonelagadda\}@kaliber.ai} \\
}

\newcommand{\Aneesh}[1]{{\color{blue}{[AJ: #1]}}}

\iclrfinalcopy % Uncomment for camera-ready version, but NOT for submission.
\begin{document}
\pagestyle{plain}

\maketitle

\begin{abstract}
    Latent world models have become increasingly popular as a method to predict and plan in latent space rather than pixel space. Recent architectures, such as LeWorldModel (LeWM), jointly train the encoder and predictor using regularization techniques like SIGReg to prevent representation collapse. Even with such regularization preventing representation collapse, we identify a new world model failure mode of \textit{physical representation laziness}, particularly noted in highly dynamic environments. For these lazy cases, the learned latent states do not collapse but nonetheless fail to represent key physical properties, causing ubiquitous downstream planning failure. To resolve this issue, we propose training-time auxiliary supervision with a lightweight "Fourier auxiliary head", which enforces physically-informed structuring of the latent space with no additional inference-time cost and can be generalized to any environment. Experimentally, we show that the auxiliary head substantially improves planning success rates in dynamic environments where the baseline LeWM exhibits physical representation laziness. It also leads to modest improvements in other environments, even when the baseline does not exhibit physical representation laziness. We further observe superior planning performance being accompanied by higher latent space correlations with key physical properties, indicating both the ability of our method to physically structure latent states and the potential planning-side benefit to the learned representation being physically structured. Finally, we ablate across various training data amounts, showing that in low-data regimes, auxiliary supervision is particularly impactful in increasing success rate. These findings support the use of our Fourier auxiliary head method to improve both overall success rate and data efficiency, while avoiding representation laziness in latent world models.
\end{abstract}

\begin{figure}[t]
    \centering
    \includegraphics[width=0.8\linewidth]{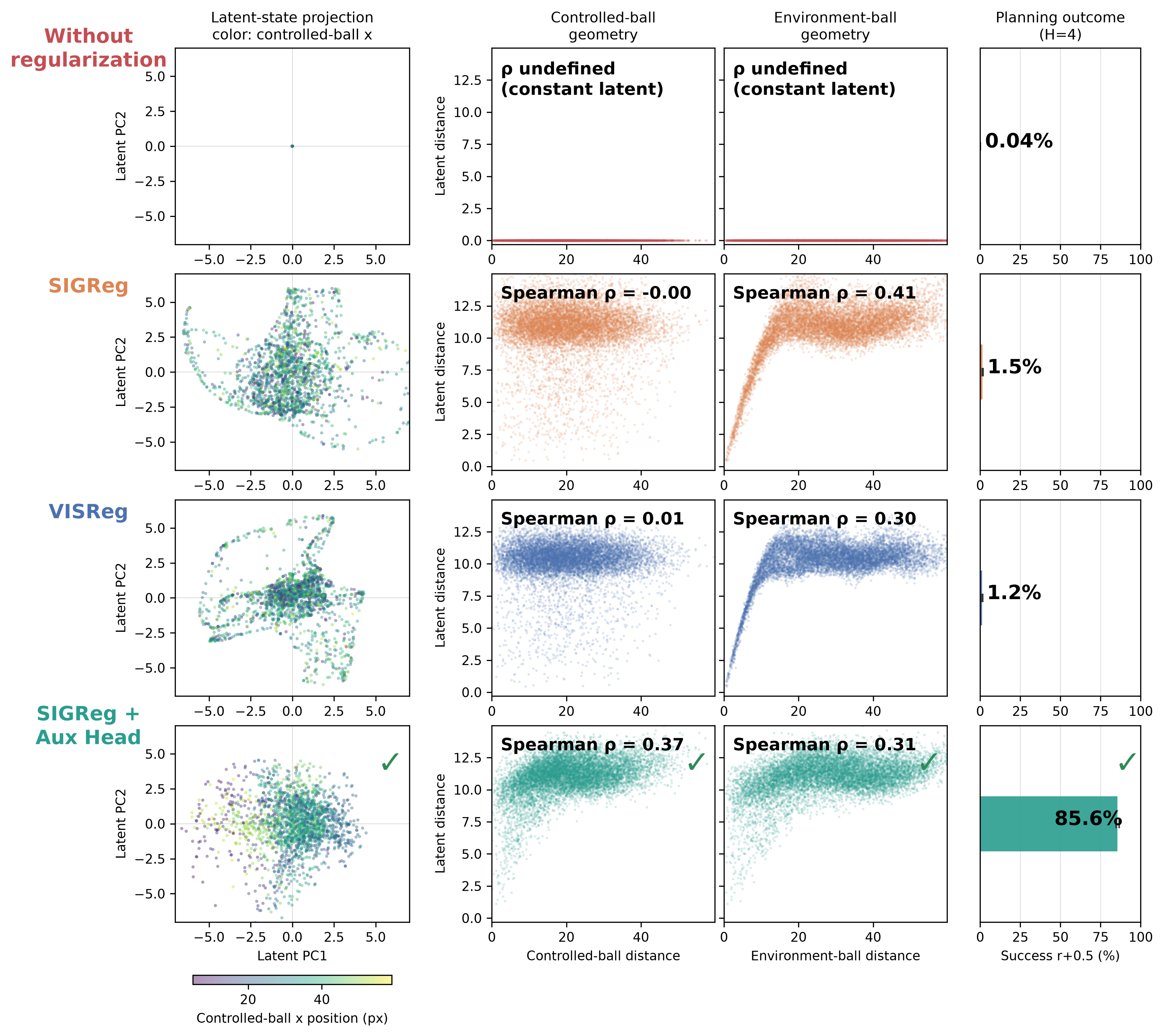}
    \caption{LeWM configurations for two-ball environment task across latent state projection (left), latent state vs position correlation (middle), and planning outcome (right). LeWM without regularization (top) leads to a collapsed representation which impedes any physical correlation or planning success. LeWM using SIGReg (second row) and VISReg (third row) both prevent representation collapse, but learn a lazy representation that only attunes to the environment ball instead of the control ball, hindering planning success. LeWM using SIGReg combined with an auxiliary head (bottom) maintains an uncollapsed representation and correlation to the control and environment ball, enabling downstream planning success.}
    \label{fig:placeholder}
\end{figure}

\section{Introduction}
\label{sec:introduction}

%\PZ{ a cover figure needed here to schematically show the ideas, I propose a three-row figure, first row demonstrate the collapse (through a latent space state plot), and the second row with SIGreg demonstrates the laziness (through the correlation plot), and the third row with both SIGreg and our aux head, which works best (with a check mark? I have not figured out a good plan for this one). }
%\Aneesh{Agreed with the general structure, I think we should also fit some success rate since interpretation of correlation and state space should probably be tied to direct results. Maybe there can be an arrow pointing downward saying "Increasing success rate" on the side of these 3 rows.}

A world model for planning should preserve the parts of a scene that matter for future actions. It does not need to reproduce every pixel, but it must represent important objects and predict how their states will change. Joint-Embedding Predictive Architectures (JEPAs) follow this idea by predicting future observations in a learned latent space rather than reconstructing them in pixel space \citep{assran2023self}. This makes the prediction more efficient and allows the model to ignore visual details that are not relevant for the underlying task.

Recent work has extended JEPAs to action-conditioned world modeling and planning. For example, LeWorldModel (LeWM) \citep{maes2026leworldmodel} jointly trains an encoder and a predictor from raw pixels using only a latent prediction loss and Sketched Isotropic Gaussian Regularization (SIGReg) \citep{balestriero2025lejepa}. The prediction loss trains the predictor to estimate the latent representation of a future observation, while SIGReg prevents the representation from collapse to a single vector by requiring the latent distribution to match an isotropic Gaussian. 

However, preventing collapse does not guarantee that the encoder preserves the physical information necessary for planning \citep{joseph2026interpreting}. There is an important difference between a \emph{collapsed} encoder and a \emph{lazy} encoder. A collapsed encoder produces nearly the same representation for every observation. A lazy encoder produces different representations, and therefore satisfies SIGReg, but captures only the minimum physical information needed to reduce the latent prediction loss. It may ignore objects or geometric relations that are important for the downstream planning task.  We demonstrate such a failure directly through a case study on bouncing-ball problems in Section~\ref{sec:motivating_experiment}. 

To address this problem, we introduce a simple auxiliary head that physically grounds the learned latent representation. During training, we assume access to the positions or bounding boxes of the objects that matter for the planning task. These annotations may come directly from a simulator or dataset, or from a lightweight object detector or tracker. The auxiliary head is a small multilayer perceptron that is trained to recover spectral features of these quantities from the encoded latent state. The same supervision can also be applied to predicted latent states, encouraging the predictor to preserve the relevant physical information over time.

This Fourier auxiliary head provides a direct learning signal that is missing from the standard JEPA objective. SIGReg asks whether the latent distribution is sufficiently diverse; the auxiliary objective asks whether this diversity contains the physical information required by the task. The two objectives therefore address different failure modes. SIGReg prevents all observations from becoming identical in latent space, while the auxiliary head prevents physically important observations from being treated as effectively identical. We provide a further theoretical discussion on the Fourier auxiliary head performance with both SIGReg and VISReg \citep{wu2026visreg}, a popular newer alternative to SIGReg, in the Appendix (\ref{app:visreg-theory}).

The proposed head is lightweight and is needed primarily during training. After training, it can be removed without changing the encoder--predictor architecture. When the planning task contains explicit rules, however, the learned head can also be retained and applied to imagined latent states. For example, predicted object positions can be used to penalize collisions, boundary violations, or other geometric constraints during planning. Thus, the auxiliary head can improve planning in two ways: indirectly, by improving the encoder and predictor, and directly, by translating clear physical rules into planning costs. In this paper, we analyze the indirect mechanism, but briefly examine direct planning improvement in the Appendix (\ref{app:aux-cem}).

Our experiments show that physical grounding through a simple auxiliary head improves the representation learned by the encoder, reduces latent prediction error, and increases downstream planning success. These improvements are particularly useful in data-limited regimes, where the standard prediction objective may not provide enough evidence for the model to discover task-relevant physical structure on its own. The auxiliary head therefore has practical value: it uses inexpensive geometric supervision to reduce the amount of interaction data needed to learn a useful world model. Overall, our results support a sharp conclusion: avoiding representation collapse is necessary, but not sufficient. A JEPA must also be guided to preserve the parts of the physical state that matter for downstream decisions.

Our main contributions are:
\begin{itemize}
    \item We identify \emph{physical representation laziness}, a failure mode in which SIGReg prevents global collapse, but the encoder still omits physical information required for prediction and planning.
    
    \item We introduce a lightweight Fourier auxiliary head that grounds encoded and predicted latent states using the positions or bounding boxes of task-relevant objects. Our permutation-aware formulation also handles visually indistinguishable objects without relying on arbitrary object identities or orderings.
    
    \item We show that auxiliary physical supervision improves the encoder, the latent predictor, and downstream planning success, with particularly useful benefits when training data are limited. When explicit physical constraints are available, the learned head can additionally support rule-aware planning.
\end{itemize}

\section{Related Work}
\label{sec:related_work}

\paragraph{World models.}
World models learn predictive dynamics for control and planning. Early approaches learn compact latent dynamics while using pixel reconstruction or prediction to train the representation \citep{ha2018worldmodels,hafner2019planet,hafner2020dreamer,micheli2023iris}. DreamerV3 follows this reconstruction-based approach and scales it across diverse domains \citep{hafner2023dreamerv3}. In contrast, TD-MPC2 learns a decoder-free latent world model using task-specific consistency, reward, and value objectives \citep{hansen2024tdmpc2}. DINO-WM also avoids pixel reconstruction, but models dynamics over features produced by a fixed pretrained DINOv2 encoder \citep{zhou2025dinowm}.

\paragraph{Joint-embedding predictive models.}
Joint-embedding predictive architectures learn by predicting target representations rather than reconstructing pixels. I-JEPA and V-JEPA demonstrate the effectiveness of this principle for images and videos \citep{assran2023ijepa,bardes2024vjepa}. In world models, PLDM trains latent dynamics using a multi-term variance--invariance--covariance objective \citep{sobal2025pldm}. LeWM instead jointly trains the encoder and predictor from pixels using a latent prediction objective and SIGReg to prevent representation collapse \citep{maes2026leworldmodel}. Very recently, there is a proposal to prevent the collapse called the Variance-Invariance-Sketching Regularization (VISReg) instead of SIGReg~\cite{wu2026visreg}. Our work builds on this end-to-end setting but studies a different failure mode: the representation can remain non-collapsed while omitting physical variables required for planning.

\paragraph{Extensions of LeWM.}
Recent work improves LeWM along several complementary directions. Fast-LeWM predicts action prefixes to accelerate long-horizon planning, Hi-LeWM introduces hierarchical latent subgoals, and AdaJEPA adapts the model online during planning \citep{gao2026fastlewm,caselli2026hilewm,wang2026adajepa}. Other studies refine latent regularization through temporally centered SIGReg or quantile-based distribution matching \citep{liu2026centeredsigreg,yu2026qqworld}. Related to \cite{auxrandall2025} which explains a theoretical motivation to auxiliary head training on LeWM, RC-aux and Temporal-Distance JEPA further introduce auxiliary objectives that make latent distances more useful for planning \citep{li2026rcaux,bai2026tdjepa}.

Most closely related to our work, depth regularization supplies geometric supervision during training, while PSG-JEPA grounds representations using robot proprioception and joint-state changes \citep{khan2026depthjepa,yan2026psgjepa}. We study task-relevant object grounding: a lightweight auxiliary head supervises the positions or bounding boxes of important objects in both encoded and predicted states. We further consider limited-data training, scenes containing visually indistinguishable objects, and planning tasks with explicit physical rules. This provides a simple way to reduce physical representation laziness while improving prediction and downstream planning.
% Test test \cite{Hinton06} test

% JEPA-style world models are revolutionary because of the existing world model archetypes, they are the computationally lightest and the most generalizable in training. This is a feature of the ability to train both the encoder that serves as the observation of state and predictor that functions as the state propagator at once, theoretically ensuring that only the minimum distilled information necessary for state propagation is captured by both the encoder and the predictor. Existing world model architectures all feature pre-trained components that will inherently contain extraneous information to the action conditioned state propagation task that unify all world models. WAM and other computationally heavy world models maintain video generation models as their backbone and utilize . Other world model methods like DINO-WM utilize pretrained encoders to capture features necessary for next state propagation, however this is heavily dependent on both the ability and reliability of the existing encoder for success in a dynamic environment. 

\section{Beyond Representation Collapse: A Case Study of Encoder Laziness in JEPA World Models}
\label{sec:BallCaseStudy}

\begin{figure}[h!]
    \centering
    \includegraphics[width=0.75\linewidth]
    {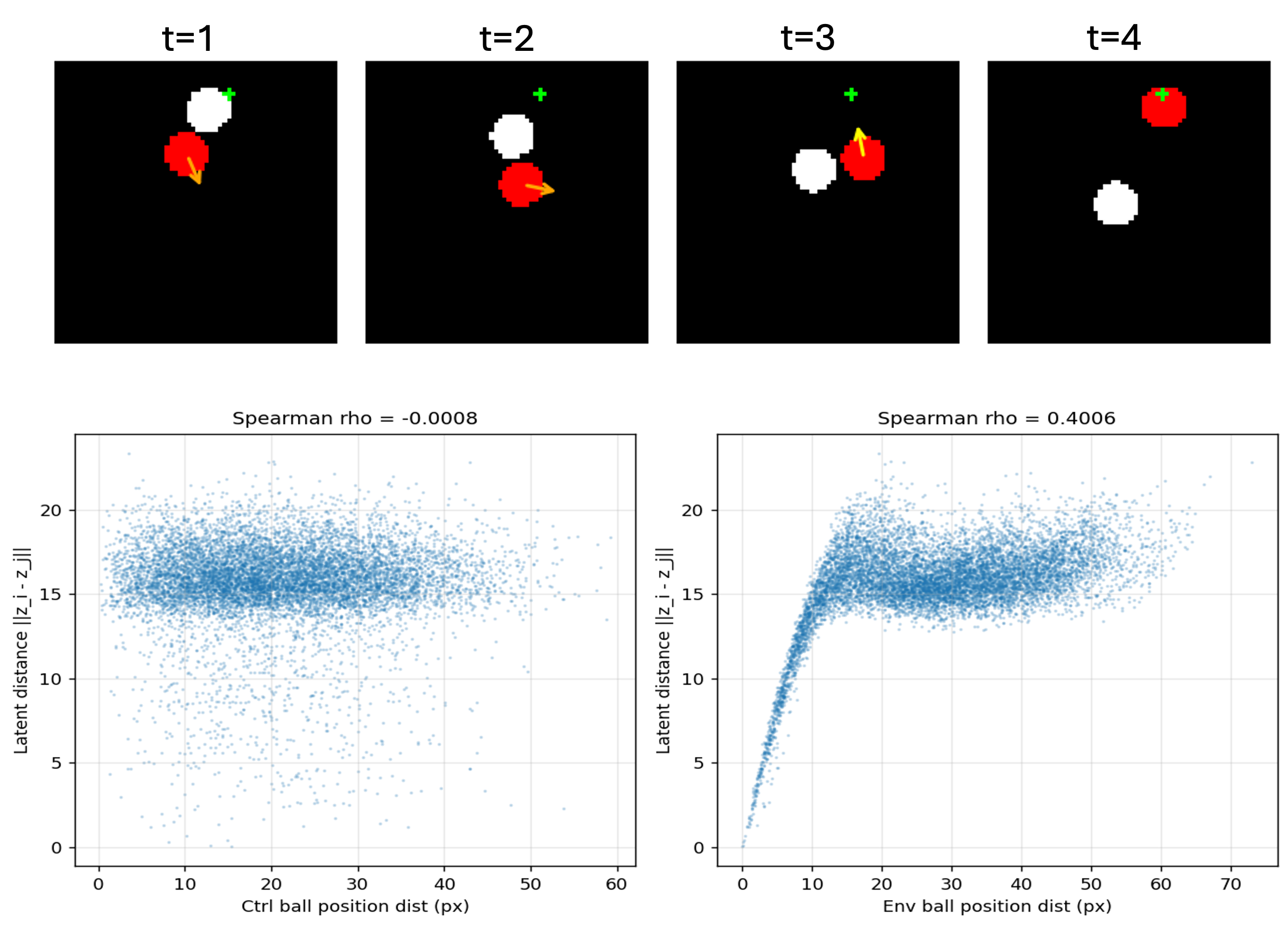}
    \caption{
    \textbf{Encoder laziness despite SIGReg.}
    Top: example frames from the two-ball task. The orange arrows indicate actions applied to the controlled ball.
    Bottom: latent distance versus the position difference of each ball. The latent space captures the environmental-ball position
    (\(\rho_{\mathrm{env}}=0.4006\)) but not the controlled-ball position
    (\(\rho_{\mathrm{ctrl}}=-0.0008\)).
    }
    \label{fig:encoder_laziness}
\end{figure}

\label{sec:motivating_experiment}

We consider a simple two-ball environment shown in Figure~\ref{fig:encoder_laziness}. The white environmental ball has an initial velocity, moves autonomously, and bounces off the walls. The red controlled ball moves only in response to external actions. The task is to move the controlled ball from its initial position to the green target without colliding with the environmental ball. One expects the world model to plan the actions needed to complete the task given the initial frame and final frame. Successful planning therefore requires the model to capture positions of both balls.

We train LeWM using its latent prediction loss and SIGReg. To measure which physical information is preserved, we sample pairs of observations \((o_i,o_j)\), encode them as \(z_i\) and \(z_j\), and compute their latent distance
\[
d_z(i,j)=\|z_i-z_j\|_2.
\]
For each ball \(b\in\{\mathrm{ctrl},\mathrm{env}\}\), we compute its position difference
\[
d_b(i,j)=\|p_i^b-p_j^b\|_2
\]
and report the Spearman correlation between \(d_z\) and \(d_b\). The nonzero and broadly distributed latent distances confirm that representation collapse does not occur. However, the near-zero correlation for the controlled ball shows that its position is not preserved in the latent geometry. The encoder is therefore non-collapsed but physically lazy: it captures the position of only one ball, although planning requires both.

This example shows that SIGReg prevents global collapse but does not determine which physical variables the encoder should preserve. Consequently, the learned latent space does not provide a reliable planning objective. Representative planning failures are included in the supplementary material. This limitation motivates the physical grounding through lightweight auxiliary head, which will be discussed in details in the following.

\section{Methods}

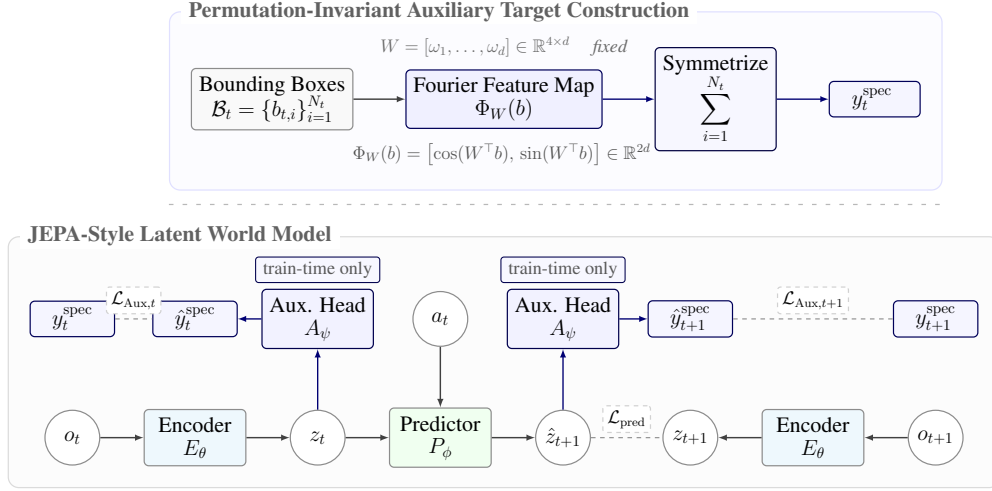
\begin{figure*}[h!]
\centering
\resizebox{0.94\textwidth}{!}{%
\begin{tikzpicture}[
    font=\Large,
    node distance=0.85cm and 1.0cm,
    block/.style={draw=black!45,line width=0.8pt,rounded corners=3pt,rectangle,align=center,minimum width=2.35cm,minimum height=1.05cm,inner sep=5pt},
    state/.style={draw=black!45,line width=0.8pt,circle,align=center,minimum width=36pt,minimum height=36pt,inner sep=1pt,fill=white},
    target/.style={draw=blue!35!black,line width=0.8pt,rounded corners=3pt,rectangle,align=center,minimum width=1.95cm,minimum height=0.82cm,inner sep=4pt,fill=blue!4},
    encoder/.style={block,fill=cyan!6},
    predictor/.style={block,fill=green!6},
    aux/.style={block,fill=blue!5,draw=blue!30!black},
    fourier/.style={block,fill=blue!6,draw=blue!35!black},
    aggregate/.style={block,fill=blue!3,draw=blue!25!black},
    flow/.style={->,>=latex,line width=0.95pt,draw=black!75},
    auxflow/.style={->,>=latex,line width=0.9pt,draw=blue!45!black},
    lossedge/.style={dashed,line width=0.8pt,draw=black!42},
    loss/.style={font=\large,fill=white,draw=black!18,rounded corners=2pt,inner sep=3pt},
    trainbox/.style={
        draw=blue!30!black,
        fill=blue!3,
        line width=0.7pt,
        rounded corners=2pt,
        font=\large,
        text=black!65,
        inner xsep=5pt,
        inner ysep=3pt,
        align=center
    },
    paneltitle/.style={font=\Large\bfseries,text=black!62,fill=white,inner xsep=5pt,inner ysep=2pt}
]
% ==================================================
% BOTTOM: LATENT WORLD MODEL
% ==================================================
\begin{scope}[local bounding box=bottom_model]
\node[state] (ot) {$o_t$};
\node[encoder,right=of ot] (enc1) {Encoder\\[-1pt]$E_\theta$};
\node[state,right=of enc1] (zt) {$z_t$};
\node[predictor,right=of zt] (pred) {Predictor\\[-1pt]$P_\phi$};
\node[state,right=of pred] (zhat_next) {$\hat z_{t+1}$};
\node[state,right=1.65cm of zhat_next] (z_next) {$z_{t+1}$};
\node[encoder,right=of z_next] (enc2) {Encoder\\[-1pt]$E_\theta$};
\node[state,right=of enc2] (o_next) {$o_{t+1}$};

\node[state,above=1.40cm of pred] (at) {$a_t$};

\node[aux,above=1.40cm of zt] (aux1) {Aux. Head\\[-1pt]$A_\psi$};
\node[trainbox,above=0.13cm of aux1] (train1) {train-time only};

\node[target] (yhat) at (aux1 -| enc1) {$\hat y_t^{\mathrm{spec}}$};
\node[target] (yt) at (aux1 -| ot) {$y_t^{\mathrm{spec}}$};

\node[aux,above=1.40cm of zhat_next] (aux2) {Aux. Head\\[-1pt]$A_\psi$};
\node[trainbox,above=0.13cm of aux2] (train2) {train-time only};

\node[target] (yhat_next) at (aux2 -| z_next) {$\hat y_{t+1}^{\mathrm{spec}}$};
\node[target] (y_next) at (aux2 -| o_next) {$y_{t+1}^{\mathrm{spec}}$};

\draw[flow] (ot) -- (enc1);
\draw[flow] (enc1) -- (zt);
\draw[flow] (zt) -- (pred);
\draw[flow] (pred) -- (zhat_next);
\draw[flow] (at) -- (pred);

\draw[flow] (o_next) -- (enc2);
\draw[flow] (enc2) -- (z_next);

\draw[auxflow] (zt) -- (aux1);
\draw[auxflow] (aux1) -- (yhat);

\draw[auxflow] (zhat_next) -- (aux2);
\draw[auxflow] (aux2) -- (yhat_next);

\draw[lossedge] (yt) -- (yhat)
    node[midway,above=3pt,loss] {$\mathcal{L}_{\mathrm{Aux},t}$};

\draw[lossedge] (zhat_next) -- (z_next)
    node[midway,above=3pt,loss] {$\mathcal{L}_{\mathrm{pred}}$};

\draw[lossedge] (yhat_next) -- (y_next)
    node[midway,above=3pt,loss] {$\mathcal{L}_{\mathrm{Aux},t+1}$};

\end{scope}

% ==================================================
% TOP: FOURIER-SYMMETRIZED TARGET CONSTRUCTION
% ==================================================
\coordinate (topcenter) at ([yshift=3.65cm]bottom_model.north);

\node[
    fourier,
    minimum width=3.8cm
] (fourier_map) at (topcenter)
{
    Fourier Feature Map\\[-1pt]
    $\Phi_W(b)$
};

\node[
    block,
    fill=gray!5,
    left=1.20cm of fourier_map,
    minimum width=3.25cm
] (bbox_set)
{
    Bounding Boxes\\[-1pt]
    $\mathcal{B}_t=\{b_{t,i}\}_{i=1}^{N_t}$
};

\node[
    aggregate,
    right=1.20cm of fourier_map,
    minimum width=2.65cm
] (sum_pool)
{
    Symmetrize\\[-1pt]
    $\displaystyle\sum_{i=1}^{N_t}$
};

\node[
    target,
    right=1.20cm of sum_pool,
    minimum width=2.10cm
] (spectral_target)
{
    $y_t^{\mathrm{spec}}$
};

\draw[flow] (bbox_set) -- (fourier_map);
\draw[auxflow] (fourier_map) -- (sum_pool);
\draw[auxflow] (sum_pool) -- (spectral_target);

\node[
    below=0.20cm of fourier_map,
    font=\large,
    align=center,
    text=black!68
] (phi_def)
{
    $\Phi_W(b)=
    \bigl[
        \cos(W^\top b),\,
        \sin(W^\top b)
    \bigr]
    \in\mathbb{R}^{2d}$
};

\node[
    above=0.17cm of fourier_map,
    font=\large,
    text=black!58
] (w_def)
{
    $W=[\omega_1,\ldots,\omega_d]
    \in\mathbb{R}^{4\times d}$
    \quad \textit{fixed}
};

% ==================================================
% BACKGROUND PANELS
% ==================================================
\begin{scope}[on background layer]

\node[
    draw=blue!12,
    fill=blue!1,
    rounded corners=7pt,
    line width=0.8pt,
    fit=(bbox_set)(fourier_map)(sum_pool)(spectral_target)(phi_def)(w_def),
    inner xsep=14pt,
    inner ysep=13pt
] (target_panel) {};

\node[
    draw=black!14,
    fill=black!1,
    rounded corners=7pt,
    line width=0.8pt,
    fit=(ot)(enc1)(zt)(pred)(zhat_next)(z_next)(enc2)(o_next)(at)(aux1)(aux2)(yt)(yhat)(yhat_next)(y_next)(train1)(train2),
    inner xsep=14pt,
    inner ysep=13pt
] (model_panel) {};

\end{scope}

% ==================================================
% PANEL LABELS
% ==================================================
\node[
    paneltitle,
    anchor=west
] at ([xshift=8pt]target_panel.north west)
{Permutation-Invariant Auxiliary Target Construction};

\node[
    paneltitle,
    anchor=west
] at ([xshift=8pt]model_panel.north west)
{JEPA-Style Latent World Model};

% ==================================================
% DIVIDER
% ==================================================
\draw[
    line width=0.8pt,
    loosely dashed,
    draw=black!28
]
([yshift=-0.32cm]target_panel.south west)
--
([yshift=-0.32cm]target_panel.south east);

\end{tikzpicture}%
}
\caption{\textbf{Overview of the proposed Fourier-symmetrized auxiliary supervision framework.}
Bounding-box states for exchangeable objects are mapped through a fixed Fourier feature map and symmetrized by summation to form the spectral target $y_t^{\mathrm{spec}}$. During training, auxiliary heads encourage both encoded and predicted latent states to preserve this geometric information. The auxiliary heads are discarded at inference time.}
\label{fig:architecture}
\end{figure*}
\subsection{JEPA-Style World Models}

JEPA-style world models, such as LeWM, are built by jointly training the entire architecture with a focus on latent representation over pixel reconstruction. They consist of an encoder ($E_{\theta}$) that maps a given frame observation $\sigma_{t}$ to latent $z_{t}$, and a predictor ($P_{\phi}$) that predicts the following latent $\hat{z}_{t+1}$ from $z_{t}$ conditioned by action $a_{t}$, i.e., 

$$z_t = E_{\theta}(o_t), \quad \hat{z}_{t+1}=P_{\phi}(z_t, a_t).$$
$$$$

Naturally, the model should be trained by a prediction loss which is the mean-squared error (MSE) of the predicted and encoded latent:

\begin{equation}
\mathcal{L}_{\text{pred}} \coloneqq ||\hat{z}_{t+1}-z_{t+1}||^2_2.
\end{equation}

% The training objective is for the predictor to learn latent representations over pixel reconstruction, like traditional world modeling. Therefore, the prediction loss is the mean-squared error of the predicted and encoded latent:

% $$
% \mathcal{L}_{\text{pred}} \coloneqq ||\hat{z}_{t+1}-z_{t+1}||^2_2.$$

This loss alone, however, leads to a lazy optimization of the latent known as representation collapse, where the encoder learns to map all observations into the same latent to minimize the prediction loss. To prevent this representation collapse, LeWM introduces the Sketched-Isotropic-Gaussian Regularizer (SIGReg)~\cite{balestriero2025lejepa,maes2026leworldmodel}. It projects embeddings into $M$ random unit-norm directions and optimizes the univariate Epps-Pulley test statistics \citep{epps1983test}, effectively enforcing a Gaussian-distributed latent embedding. The complete training loss, with hyperparameter $\lambda_{S}$ is thus as follows: 

\begin{equation}
    \mathcal{L}_{\text{LeWM}} \coloneqq \mathcal{L}_{\text{pred}} + \lambda_{S} \text{SIGReg}(Z)
\end{equation}

More recently, VISReg~\citep{wu2026visreg} extends sketching-based methods to be flexible to distribution shape, and can been used a drop-in replacement for SIGReg. Although both SIGReg and VISReg help prevent representation collapse, they do not guarantee that the learned latent representation preserves all the key information required for planning. To address this limitation, we introduce an auxiliary prediction head that infers key geometric properties of predefined, task-relevant objects directly from the latent state. We choose to use SIGReg for our method as opposed to VISReg, for reasons empirically and theoretically examined in Appendix \ref{app:visreg-theory}.

\subsection{Fourier-Symmetrized Bounding Box Auxiliary Head }

We use the bounding boxes of objects to represent structural information. This serves as a sparse and visually obtainable representation of object geometry, entirely in pixel space,  which also analogously preserves depth information in 3D environments. Bounding boxes can also be calculated using foundation object-detection models in order to construct a ground-truth target from real visual data; as a result our method imposes little-to-no additional manual annotation or supervision footprint. 

We thus define the geometric representation of the environment at time $t$ as the set of bounding-box states $\mathcal{B}_t = \{b_{t,i}\}_{i=1}^{N^\text{id}_t}$, constructed from the bounding boxes $N^\text{id}_t$ individual objects. For the $i$-th object at time $t$, each state vector $\mathbf{b}_{t,i}$ is defined as:
%$$\mathbf{b}_{t,i} = [x_{t,i},y_{t,i},w_{t,i},h_{t,i},\sin(\theta_{t,i}),\cos(\theta_{t,i}),COM_{x,t,i},COM_{y,t,i}] \in \mathbb{R}^8$$
$$\mathbf{b}_{t,i} = [x_{t,i},y_{t,i},w_{t,i},h_{t,i}] \in \mathbb{R}^4$$

where $(x_{t,i},y_{t,i})$ denotes the box center, and $(w_{t,i},h_{t,i})$ its width and height respectively, in pixel coordinates. 
When all objects are visually distinct, this state vector can in theory simply be aggregated across objects via concatenation and used as the auxiliary target. However, when there are visually similar or identical objects, the objective becomes poorly defined since arbitrarily-assigned object identity affects the aggregate state vector. To address this, we instead symmetrically aggregate objects via a Fourier sum to construct a permutation-invariant spectral representation of the bounding boxes. We first define the distribution of bounding box spatial states  using point measures, for the bounding boxes of a set of visually identical objects at time $t$, i.e., $\mathcal{B}^{\text{id}}_t = \{\mathbf{b}_{t,1},...,\mathbf{b}_{t,N_t^{\text{id}}}\}$:
$$\mu_t(\mathbf{b}) =  \sum_{i=1}^{N^{\text{id}}_t}\delta(\mathbf{b}-\mathbf{b}_{t,i})$$  
where $\delta$ is the Dirac delta function~\footnote{Dirac delta function satisfies the sifting property $\int_{\mathbb{R}^{d_\mathbf{b}}} f(\mathbf{b})\,\delta(\mathbf{b}-\mathbf{b}_{t,i})\,d\mathbf{b}=f(\mathbf{b}_{t,i})$ for all continuous compactly supported functions $f$.}. $\mu_t(\mathbf{b})$ is the fundamental permutation-invariant function of $\mathbf{b}_{t,i}$'s and encodes the spatial configuration of the visually identical objects. Its Fourier-component associated with a spatial-frequency $\boldsymbol{\omega}$ is, 
$$ \hat{\mu}_t(\boldsymbol{\omega}) = \int_{\mathbb{R}^8}e^{i\boldsymbol{\omega}\cdot\mathbf{ b}}\mu_t(\mathbf{b})d\mathbf{b} =\sum_{i=1}^{N^{\text{id}}_t} e^{i\boldsymbol{\omega}\cdot \mathbf{b}_{t,i}} =  \sum_{i=1}^{N^{\text{id}}_t}\cos(\boldsymbol{\omega}\cdot \mathbf{b}_{t,i}) + i \sin(\boldsymbol{\omega}\cdot \mathbf{b}_{t,i}), $$

and captures detailed aspects of the spatial configuration while inheriting the permutation invariance of $\mu_{t}(\mathbf{b})$. To obtain a discrete target, we sample this continuous function over a set of $d$ Fourier frequencies by generating a frequency matrix $W = [\omega_1, ... , \omega_d] \in \mathbb{R}^{4\times d}$ where each column $\omega_k$ represents a sampled frequency. We define the symmetrized bounding box target as:

\begin{equation}
   y_t^{\text{spec}} := [\mathrm{Re}( \hat{\mu}_t(W)),\mathrm{Im}( \hat{\mu}_t(W))] =\sum_{i=1}^{N^{\text{id}}_t}\Phi_{W}(\mathbf{b}_{t,i}),\quad \Phi_{W}(\mathbf{b})\:=\left[\cos(W^{\top}\mathbf{b}),
\sin(W^{\top}\mathbf{b})\right]  
\end{equation}
 %\left[
%\operatorname{Re}\hat{\mu}_t(\omega_1),\ldots,
%\operatorname{Re}\hat{\mu}_t(\omega_d),
%\operatorname{Im}\hat{\mu}_t(\omega_1),\ldots,
%\operatorname{Im}\hat{\mu}_t(\omega_d)
%\right].

%  Equivalently using the Euler identity we define the target $y_t^{\text{spec}}$ and deterministic Fourier feature map $\Phi_W(b)$ as
%  $$
% y_t^{\text{spec}} := \sum_{i=1}^{N_t}\left[\cos(W^{\top}b_{t,i}),
% \sin(W^{\top}b_{t,i})\right] \in \mathbb{R}^{2d} \quad \Phi_W(b):=\left[\cos(W^\top b),\sin(W^\top b)\right]
% $$

%While a simple target of the sum of state vectors $b_{t,i}$ is permutation-invariant, it suffers from lack of uniqueness. Our method instead leverages 
The uniqueness theorem for Fourier transforms states that a measure is uniquely determined by its Fourier transform, i.e., $\mu_1 = \mu_2 \iff \hat{\mu}_1(\omega) = \hat{\mu}_2(\omega)\ \forall \omega \in \mathbb{R}^d$. Consequently, two different bounding box configurations for any set of objects, regardless of distinguishability, will differ by at least one Fourier component. Our method samples a finite set of frequencies so the resulting representation $y^{\text{spec}}_t$ is not necessarily injective, but increasing $d$ yields a richer approximation of the full Fourier transform representation, reducing degeneracy among distinct configurations. As opposed to a simple pixel-space object location sum, which is also a permutation-invariant target, the Fourier target exhibits high, almost guaranteed, uniqueness.   

\subsection{Architecture of the Auxiliary head}

The auxiliary head $A_{\psi}$ is designed as a simple 2-layer MLP that predicts the spectral targets from the latents:

$$\hat{y}^{\text{spec}}_t=A_{\psi}(z_t),$$ %\qquad \hat{y}^{\text{spec}}_{t+1}=A_{\psi}(\hat{z}_{t+1})$$

which leads to an auxiliary loss for each latent state $z_t$:

\begin{equation}
    \mathcal{L}_{\text{Aux},t} \coloneqq ||y^{\text{spec}}_t-\hat{y}^{\text{spec}}_t||^2_2.
\end{equation}

%\qquad \mathcal{L}_{\text{Aux}, t+1} \coloneqq ||y^{\text{spec}}_{t+1}-\hat{y}^{\text{spec}}_{t+1}||^2_2$$

Propagation of the auxiliary losses results in separate loss terms for the auxiliary head, encoder, and predictor. The auxiliary head is trained only on the loss term $\mathcal{L}_{Aux,t}$, as predicting the spectral target from the predicted latent could result in instability. Hyperparameters $\lambda_{A1}$ and $\lambda_{A2}$ are used to integrate the auxiliary losses.

%\mathcal{L}_{A_{\psi}}=\mathcal{L}_{\text{Aux},t}
%\qquad
\begin{equation}
\mathcal{L}_{P_{\phi}}=\mathcal{L}_{\text{pred}}+\lambda_{A2}\mathcal{L}_{\text{Aux},t+1},
\quad
\mathcal{L}_{E_{\theta}}=\mathcal{L}_{P_{\phi}}+\lambda_{A1}\mathcal{L}_{\text{Aux},t}+\lambda_{S}\text{SIGReg}(Z)    
\end{equation}

The auxiliary heads are thus used to further optimize the encoder and predictor during training time, and are removed during inference. We leave the auxiliary head as shallow as possible to offload learning to the main world model, as we notice laziness onset with slightly deeper architectures. 

%Aneesh Comment: This can probably just go in the results or appendix or wherever as discussed in slack. Just commenting this out for now$
\begin{comment}
\subsubsection{Hungarian Matching}

Naively predicting the state from the latent is intractable for a key reason: the predicted states for visually indistinguishable objects may be in a different order than the environment's true state vector. To bridge this gap, the Hungarian algorithm is used on the set of $N$ objects to find a permutation $\sigma \in \mathfrak{S}_N$ that minimizes cost:

$$\hat{\sigma}=\arg\min_{\sigma \in \mathfrak{S}_N} \sum_{i}^{N} \mathcal{L}_{match}(s_{i,t}, \hat{s}_{\sigma(i,t)})$$

Where $\mathcal{L}_{match}$ is a pair-wise matching cost calculated similarly to $\mathcal{L}_{Aux_{t}}$, and the $p_{obj}$ variable is used to mask objects that are not present in the scene.  
\end{comment}

\subsection{Trajectory Planning}
\label{sec:TrajectoryPlanning}

To evaluate the performance of the models, the Cross-Entropy Method (CEM) \citep{rubinstein2004cross} is used to optimize trajectories in the world model latent space, identically to the method utilized in LeWM. An initial observation $o_1$ and a randomly sampled candidate action sequence $a_{1:H}$ are used to roll out a series of latent states using the predictor, up to the planning horizon $H$. CEM iteratively samples action sequences by selecting the best plans and using their statistics to update the sampling distribution. This is used to optimize the action sequence to minimize the objective of a terminal latent goal, as shown in the equations below.

$$C(\hat{z}_H)=||\hat{z}_H-z_g||^2_2, \quad a^{*}_{1:H}=\arg\min_{a_{1:H}}C(\hat{z}_H)$$

For certain environments where clear rules are provided for the planning, the auxiliary head can also be used during inference to predict geometric properties that can augment this planning objective, which we examine in the Appendix \ref{app:aux-cem}.

\begin{comment}
\begin{figure}[h!]
    \centering
    \includegraphics[width=0.75\linewidth]
    {plan.png}
    \caption{
    \textbf{Planning and Auxhead in planning.}
    }
    \label{fig:encoder_laziness}
\end{figure}
\end{comment}

\section{Results}
\subsection{Evaluation Setup}
We tested a variety of environments with the auxiliary head architecture to ensure its robustness across tasks and settings. The environments included 1,3, and 6 environment ball(s) versions of the ball environment (100k training) described in Section \ref{sec:BallCaseStudy} as well as Push-T (500k training), Two Room (730k training), and OGBench Cube (500k training).
\subsubsection{Evaluation Metrics}

\begin{itemize}
    \item \textbf{Planning Success Rate.} As described in section \ref{sec:TrajectoryPlanning}, a standardized planning protocol is used to optimize the trajectories using each trained model in the exact same evaluation data set. The success rates of the trajectories across environments demonstrate the utility of the auxiliary heads in training for latent-space planning. We perform planning on 5 separate seeds and average the results in the tables below.

    \item \textbf{Latent Distance Correlations.} To evaluate the ability of the latent space to effectively represent object locations and velocities, we sample 2000 observations and create pairs $o_1, o_2$, obtain their states $s_1, s_2$, and encode them into their latent representations $z_1, z_2$. For these pairs, we compare the difference in states between observations and the distances between the latents, and calculate the Spearman rank correlation coefficient $\rho$ to quantify the correlation. A higher $\rho$ indicates a greater ability of the latent space to represent the object states.
\end{itemize}

\subsection{Planning Success}

To evaluate the direct benefit of the auxiliary head architecture in task completion, we report  planning success using the methodology described in section \ref{sec:TrajectoryPlanning}.

\begin{table}[h!]
\centering
\caption{Planning success rates for the n-ball environments at various prediction horizons. Best results are in bold.}
\label{tab:horizon_success_nball}
\begin{tabular}{l cc cc cc}
\toprule
 & \multicolumn{2}{c}{$H=4$} & \multicolumn{2}{c}{$H=6$} & \multicolumn{2}{c}{$H=8$} \\
\cmidrule(lr){2-3} \cmidrule(lr){4-5} \cmidrule(lr){6-7}
\textbf{Task} & \textbf{Ours} & \textbf{LeWM} & \textbf{Ours} & \textbf{LeWM} & \textbf{Ours} & \textbf{LeWM} \\
\midrule
Bouncing Balls ($n=1$) & \textbf{92.4\%} & 1.1\% & \textbf{78.7\%} & 0.2\% & \textbf{62.1\%} & 0.0\% \\
Bouncing Balls ($n=3$) & \textbf{81.2\%} & 2.7\% & \textbf{59.4\%} & 0.8\% & \textbf{29.4\%} & 0.0\% \\
Bouncing Balls ($n=6$) & \textbf{67.2\%} & 1.9\% & \textbf{43.7\%} & 0.1\% & \textbf{13.0\%} & 0.4\% \\
\bottomrule
\end{tabular}
\end{table}

\begin{table}[h!]
\centering
\caption{Spearman $\rho$ correlations between latent and pixel space in n-ball environments.}
\label{tab:rho_nball}
\begin{tabular}{l cc cc cc}
\toprule
& \multicolumn{2}{c}{Control Ball} & \multicolumn{2}{c}{Env. Ball} & \multicolumn{2}{c}{All Balls} \\
\cmidrule(lr){2-3} \cmidrule(lr){4-5} \cmidrule(lr){6-7}
\textbf{Task} & \textbf{Ours} & \textbf{LeWM} & \textbf{Ours} & \textbf{LeWM} & \textbf{Ours} & \textbf{LeWM} \\
\midrule
Bouncing Balls ($n=1$)
& \textbf{0.431} & -0.010
& 0.325 & \textbf{0.423}
& \textbf{0.506} & 0.247 \\

Bouncing Balls ($n=3$)
& \textbf{0.405} & -0.004
& 0.336 & \textbf{0.417}
& \textbf{0.476} & 0.351 \\

Bouncing Balls ($n=6$)
& \textbf{0.380} & 0.017
& \textbf{0.275} & 0.004
& \textbf{0.377} & 0.007 \\
\bottomrule
\end{tabular}
\end{table}

The results indicate improvements in planning success rates across all environments from the LeWM base formulation. This is particularly evident in our n-ball environment in Table \ref{tab:horizon_success_nball}, where LeWM has the physical representation laziness across any number of balls. Table \ref{tab:rho_nball} demonstrates poor control ball positional understanding in the latent space regardless of the number of balls, which the addition of the auxiliary target fixes by better organizing the latent space. This comes at the cost of slightly worse environment ball positional understanding for $n=1$ and $n=3$, where the default LeWM is able to capture their positions, but overall the correlations are greatly improved across all balls.

\begin{table}[h!]
\centering
\caption{Planning success rates and Spearman $\rho$ correlations on other environments.}
\label{tab:horizon_success}
\begin{tabular}{l cc cc}
\toprule
& \multicolumn{2}{c}{Success Rate} & \multicolumn{2}{c}{Spearman $\rho$} \\
\cmidrule(lr){2-3} \cmidrule(lr){4-5}
\textbf{Task} & \textbf{Ours} & \textbf{LeWM} & \textbf{Ours} & \textbf{LeWM} \\
\midrule
Push-T
& \textbf{95.2\%} & 91.2\%
& \textbf{0.144} & 0.138 \\

Two Room
& \textbf{86.0\%} & 84.0\%
& \textbf{0.425} & 0.410 \\

Reacher
& \textbf{89.6\%} & 84.8\%
& 0.008 & \textbf{0.010} \\

OG-Bench Cube
& \textbf{70.0\%} & 66.4\%
& 0.003 & \textbf{0.020} \\
\bottomrule
\end{tabular}
\end{table}

In other environments, shown in Table \ref{tab:horizon_success}, more modest but consistent increases in planning performance are reported, demonstrating the utility of this method in both 2D and 3D environments. Spearman correlation is also improved for Push-T and Two Room, though the latent spaces collapse for Reacher and OG-Bench Cube despite having improved success rates, potentially indicating the ability of the auxiliary heads to continue improving the latent spaces in other useful ways, or the model simply being capable of 'brute-forcing' the task learning without strong intrinsic physical representation. The latter is corroborated by strong correlations for Reacher and OG-Bench Cube at lower data amounts (even 500k). 

\subsection{Data Efficiency}
\begin{figure}[h]
    \centering
    \includegraphics[width=0.7\linewidth]{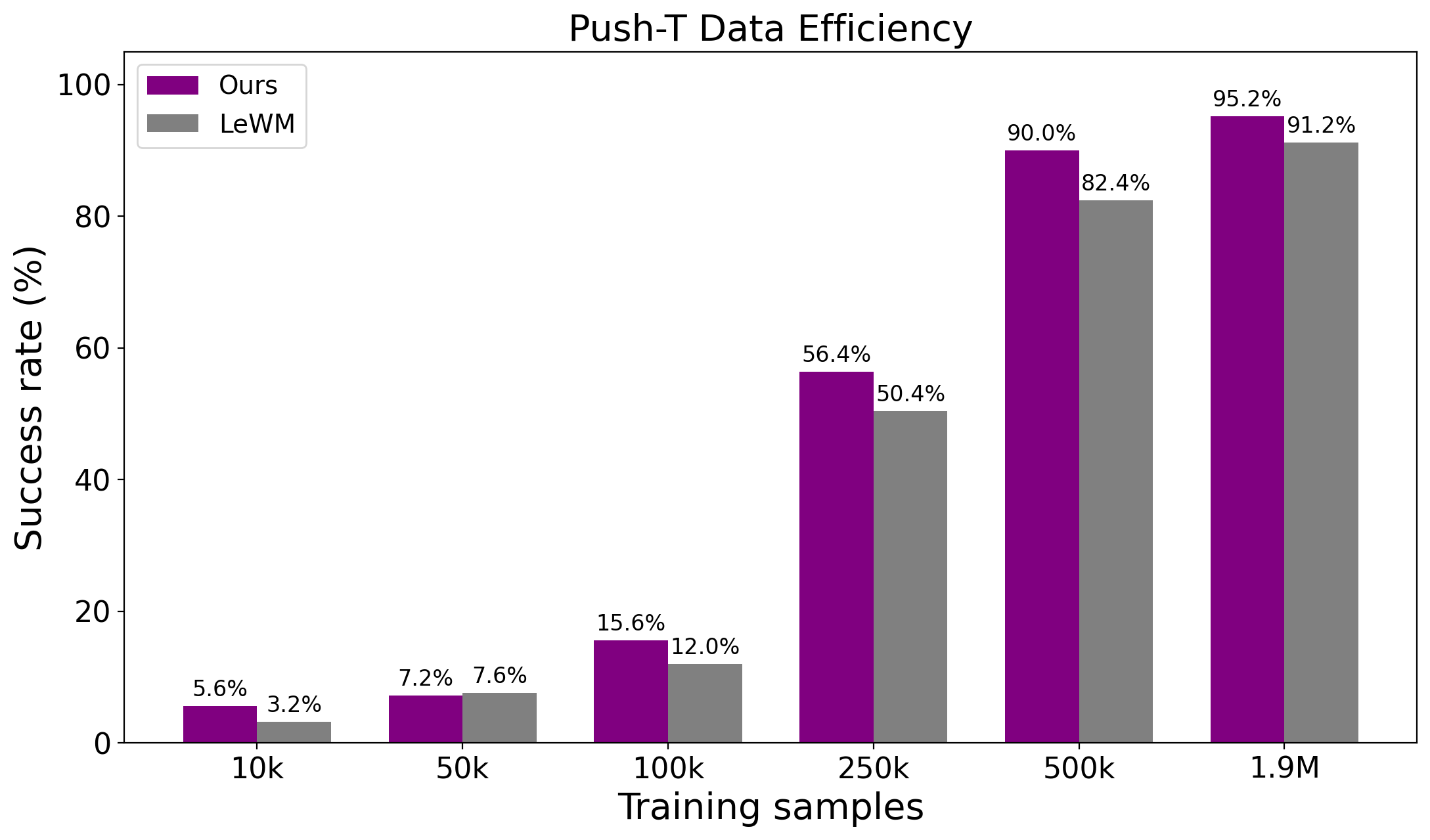}
    \caption{Push-T success rate across various training data amounts. The Fourier auxiliary head yields improvements in success rates compared to LeWorldModel in lower data regimes.}
    \label{fig:data-efficiency}
\end{figure}

\begin{comment}
Main figure: Y-axis Success rate vs X axis training data size (in episodes), two lines on figure: 1 for LeWM (dotted), 1 for our method (solid)
Environments (Reacher, PushT, OGBench Cube)
\end{comment}
One of the dimensions across which we ablate the auxiliary head method is the training data. For the purposes of direct comparison with LeWM, we trained both architectures with nested datasets of various sizes to identify how our technique improves results at lower data regimes. The Push-T results in Figure \ref{fig:data-efficiency} demonstrate that the performance improvements of the auxiliary head sustain at smaller data sizes. In real-world scenarios where data is limited, this method can still provide benefit with no inference time cost. However, it should be noted that if the data is low enough, the auxiliary target is unable to improve the baseline results due to overall underdetermination.

\section{Conclusion}

We introduced Fourier auxiliary heads for JEPA-style world models, a training time supervision method that organizes the latent space for better object-centric understanding of the environment without any additional inference cost. Despite SIGReg, we first show that JEPA-based encoders often fail to properly capture physical properties of key objects in the environments, then introduced a Fourier-symmetrized target derived from  bounding boxes, making it generalizable to both 2D and 3D environments. We compared our method with the n-balls environment, where the laziness of the encoder in the LeWM baseline leads to poor planning success, while our Fourier auxiliary head yielded an \textit{increase} of planning success rate by 91.3\%, 78.5\% and 65.3\% (H=4) for 1, 3 and 6 environmental balls, respectively. Further analyzing correlations of latents with pixel-space physical properties, LeWM led to representation laziness with near-0 correlations of the control ball, while our method successfully enabled the latent space to adequately correspond to the ball position. We additionally tested our method on more common environments, where it consistently demonstrated improved task performance at full and low-data scenarios and improved latent space encoding of physical properties in certain benchmarks. While we provide a generalized formulation for the spectral target, we do not test our method on real-life environments. Additionally, we have only tested on LeWM as a baseline, whereas this method can be implemented with any JEPA-style world model. We leave examining the Fourier auxiliary head's combinations with additional supervision, regularization, and architectures, to future works.

% \section{Reproducibility Statement}

\bibliography{iclr2026_conference}
\bibliographystyle{iclr2026_conference}

\newpage
\appendix
\section{Appendix}

\subsection{Training and Evaluation Pipeline Configuration}

There are two variations of the architecture: one used for the n-balls environment, and another used for the remaining environments, which is kept as close to the original LeWM implementation as possible.

\subsubsection{N-Balls Architecture}

Rendered RGB frames are input at $64\times64$ resolution, and normalized between $[-1, 1]$. A batch size of 64 is used, and all models are trained for 200 epochs. 4 sub-trajectories are used for training, with 4 observations and 4 actions each. No frameskip is used.

\textbf{Encoder Architecture}. A small custom vision transformer is used, with a patch size of 8, token width of 64, and 4 transformer blocks with 4 attention heads. The feedforward networks (FFNs) used are 256-D, with one learnable CLS token that is projected to a 128-D output latent.

\textbf{Predictor Architecture}. The predictor is implemented as a small, action-conditioned causal transformer. It takes in 3 history frames as input, using a 128-D latent that is projected to a dimension of 64. Like the encoder, it uses 4 transformer blocks with 4 attention heads and 256-D FFNs, but includes Adaptive Layer Norm (AdaLN) \cite{xu2019understandingimprovinglayernormalization} for action conditioning. It also uses causal masking to prevent future frames from attending to the previous frame. The output is projected back to 128 dimensions.

\textbf{Auxiliary Head Architecture}. The auxiliary heads are implemented as simple, 2-layer MLPs, where the first layer projects the 128-D latent into a 64-D feature, and the second layer outputs the 18-D spectral target.  We set $\lambda_{\text{Aux}}=0.1$, and set $\lambda_{\text{SIGReg}}=0.09$, matching the value used in both official SIGReg and VISReg repositories.

\textbf{Planning Protocol}. We use the Cross-Entropy Method for planning. We sample 100 candidate action sequences, retain 10 elites per iteration, and run it for 20 iterations. The standard deviation of the action selection is set to 2.0. 200 episodes are evaluated per seed. Additionally, the norm of the actions are clipped to 3.0, limiting the speed of the control ball. The planning horizons are set to 4, 6, and 8 for the experiments, with no replanning. The initial states are set to exactly 4, 6, or 8 steps behind a target respectively, and thus success is defined as when the control ball is within a 1-pixel radius of the true target position. 

\subsubsection{LeWM Architecture}

For the other environments, the architecture closely resembles LeWM. Rendered RGB frames are input at $224\times224$ resolution, and normalized between $[0, 1]$. A batch size of 128 is used, and all models are trained for 10 epochs. A frameskip of 5 is used, meaning 5 actions are predicted for every frame. 4 sub-trajectories are used for training, with 4 observations and 4 blocks of 5 actions each. 

\textbf{Encoder Architecture}. A Vision Transformer Tiny (ViT-Tiny) model from Hugging Face is used with a patch size of 14 for the encoder. The 192-D CLS token is used as the latent. 

\textbf{Predictor Architecture}. A Vision Transformer Small (ViT-Small) model from Hugging Face is used as backbone for the predictor. Causal masking and learned position embeddings are used to accommodate a history length of 3. 

\textbf{Auxiliary Head Architecture}. The auxiliary heads are  implemented as 2-layer MLPs, where the first layer projects the 192-D latent into a 64-D feature, and the second layer outputs the 18-D spectral target. We set $\lambda_{\text{Aux}}=0.1$, and set $\lambda_{\text{SIGReg}}=0.09$, matching the value used in both official SIGReg and VISReg repositories.

\textbf{Planning Protocol}. Identically to LeWM, we use the Cross-Entropy Method for planning. We sample 300 candidate action sequences, retain 30 elites per iteration, and run it for 30 iterations. A variance scale of 1.0 is used for action sampling. 50 scenes are evaluated per seed, and the standard LeWM action scaler is used to bind actions. Model Predictive Control (MPC) is conducted using CEM, with replanning every 5 blocks of 5 actions, consisting of 25 actions altogether. A total action budget of 50 steps is given for each environment, though they can be solved in only 25 steps.

\subsubsection{Target Creation}
\label{app:target_creation}

\begin{figure}[h!]
    \centering
    \includegraphics[width=0.8\linewidth]{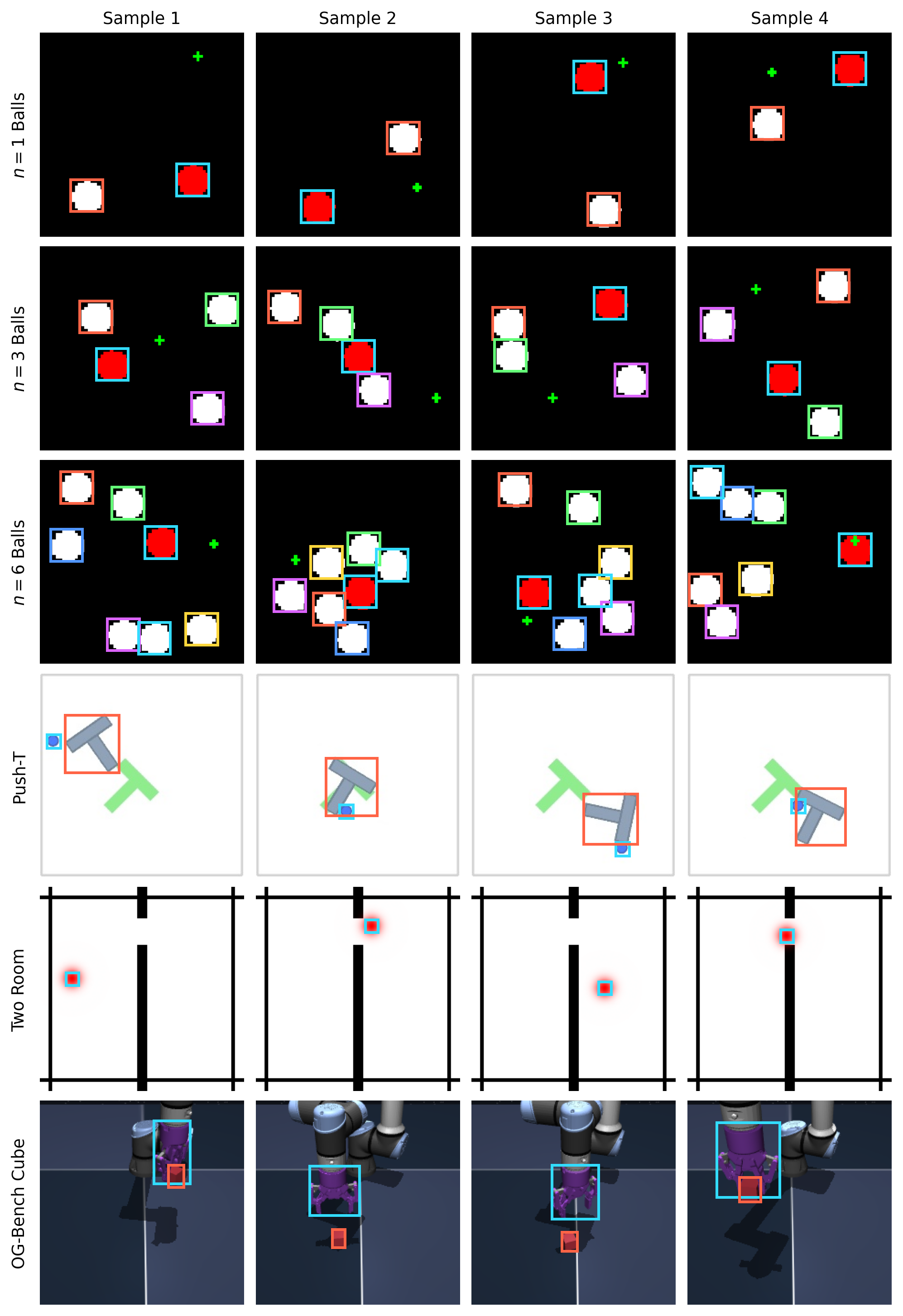}
    \caption{Sample bounding box targets for each tested environment.}
    \label{fig:obb}
\end{figure}

The bounding box targets are well-defined for each environment. For the n-ball environments, all the balls are used as targets. For the remaining environments, only dynamic objects are selected. These bounding boxes are not selected with zero-shot methods, but rather obtained as ground-truth from the environment simulators, demonstrating an ideal use-case of auxiliary supervision. Samples for each environment can be seen in Figure \ref{fig:obb}.

\begin{comment}
\textbf{Center of Mass Calculation}. To calculate the center of mass for object $i$, we take its semantic segmentation mask $M_{t,i}$, and calculate its $x$ and $y$ components as follows:

$$COM_x
=\frac{1}{|M_{t,i}|}
\sum_{(u,v)\in M_{t,i}}(u+0.5),
\qquad
COM_y
=\frac{1}{|M_{t,i}|}
\sum_{(u,v)\in M_{t,i}}(v+0.5)$$
\end{comment}

\newpage
\subsection{Theoretical Examination of VISReg + Aux Head}
\label{app:visreg-theory}

\begin{figure}[h]
    \centering
    \includegraphics[width=0.8\linewidth]{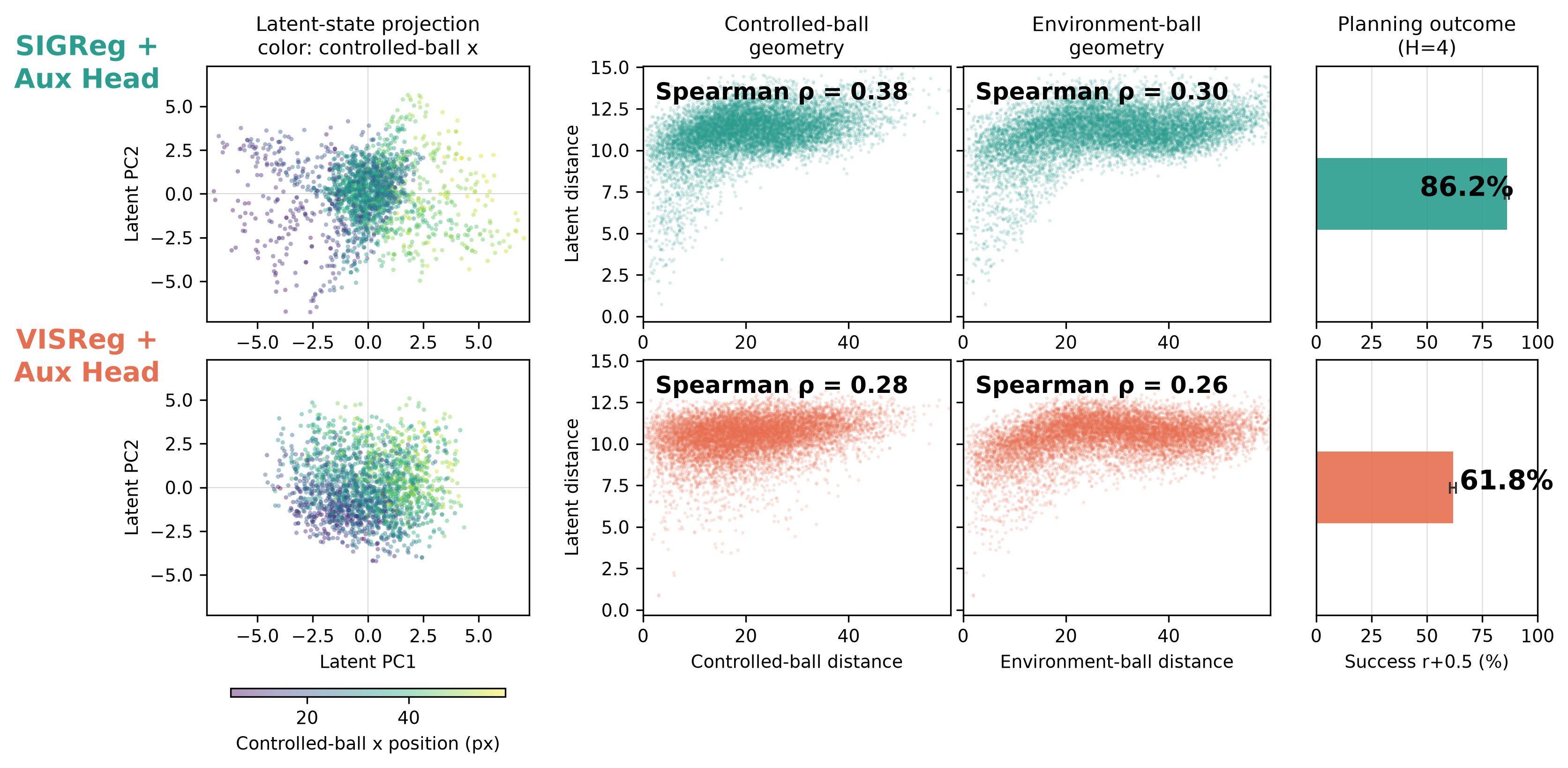}
    \caption{LeWM configurations for two-ball environment task across latent state projection (left), latent state vs position correlation (middle), and planning outcome (right). Our auxiliary head method performs worse with VISReg than with SIGReg.}
    \label{fig:appendix-2row}
\end{figure}

Interestingly, our Fourier auxiliary head training empirically yields improved planning performance with SIGReg \citep{balestriero2025lejepa} instead of VISReg \citep{wu2026visreg}, even though VISReg exhibits higher performance than SIGReg on the standard LeWorldModel. This implies that our Fourier method's benefits are not necessarily entirely additive with those of VISReg's. To examine why, we explore the theoretical relationship between VISReg and our Fourier method. VISReg utilizes the Sliced Wasserstein Distance (SWD) \citep{kolouri2019swd} which is reliant on Radon transform injectivity via the Cram\'er--Wold Theorem:
\newtheorem{lemma}{Lemma}
\begin{lemma}[Cram\'er--Wold Theorem]
Let $\mu$ and $\nu$ be probability measures on $\mathbb{R}^d$, and let
\[
\mathcal{R}\mu(\theta,t)
=
\int_{\mathbb{R}^d}
\delta\!\left(t-\langle x,\theta\rangle\right)\,d\mu(x)
\]
denote the Radon transform for $\theta \in \mathbb{S}^{d-1}$. Then $\mathcal{R}$ is injective, and
$$
\mu = \nu \quad\Longleftrightarrow \quad \mathcal{R}\mu(\theta,\cdot) = \mathcal{R}\nu(\theta,\cdot) \quad
\forall \theta \in \mathbb{S}^{d-1}. $$
\end{lemma}

This is utilized in VISReg via sampling 1D directions $\theta_k$, and attempting:
$$
\forall w_k : \mathcal{R}\mu_z(w_k,\cdot) \rightarrow \mathcal{N}(0,1)
$$
This means that across randomly sampled directions $w_k$, VISReg takes the Radon projection $\mathcal{R}\mu_z(w_k,\cdot)$ along these directions and attempts to enforce Gaussian shape and unit variance in the zero-loss limit. Since the Radon transform is the full collection $\mathcal{R}\mu_z(\theta,t)$ across all directions, VISReg stochastically regularizes the latent distribution's Radon transform. 

To connect to how this regularization principle relates to our Fourier method, the Radon transform is in fact related to the Fourier transform via the Fourier slice theorem:
\begin{lemma}[Fourier Slice Theorem]
Let $f \in L^1(\mathbb{R}^d)$, and let
\[
\mathcal{R}f(\theta,t)
=
\int_{\mathbb{R}^d}
f(x)\,
\delta\!\left(t-\langle x,\theta\rangle\right)\,dx
\]
denote its Radon transform for $\theta \in \mathbb{S}^{d-1}$.
Then the one-dimensional Fourier transform of the Radon projection
$\mathcal{R}f(\theta,\cdot)$ satisfies
\[
\mathcal{F}_t
\left[
\mathcal{R}f(\theta,t)
\right](\omega)
=
\hat{f}(\omega \theta),
\qquad
\omega \in \mathbb{R},
\]
where
\[
\hat{f}(\xi)
=
\int_{\mathbb{R}^d}
f(x)e^{-i\langle \xi,x\rangle}\,dx
\]
is the $d$-dimensional Fourier transform of $f$.
Equivalently,
\[
\mathcal{R}f(\theta,t)
=
\mathcal{F}^{-1}_{\omega}
\left[
\hat{f}(\omega\theta)
\right](t).
\]
\end{lemma}

To get a 1D Radon projection, we take a radial slice through the Fourier transform and take a 1D inverse Fourier transform to bring it back to real space. VISReg samples individual Radon projections, equivalent to these radial slices inverse-transformed, and enforces each to have unit variance and be Gaussian. This relates to our method since VISReg is essentially trying to make the Fourier Transform of the latent distribution isotropic via radial symmetry. Our method, in contrast, targets spectral \textit{anisotropy}, reflective of the spatial distribution, and supervises the latent to reflect this anisotropy. 

$$
\begin{aligned}
\text{VISReg in Radon space}
&\;\xleftrightarrow{\text{Fourier Slice Theorem}}\;&
\text{isotropy in spectral space}
\\[6pt]
\text{Aux. supervision in spatial state-space}
&\;\xleftrightarrow{\text{Fourier-feature Transform}}\;&
\text{anisotropy in spectral space}
\end{aligned}
$$

We therefore hypothesize that our method works better with SIGReg since the SIGReg regularization gradient tends to be weak for low variance regimes, whereas VISReg is explicitly designed to impose a strong corrective gradient along sampled low-variance directions. As a result, SIGReg's weak gradient corrections thus may permit task-induced latent anisotropy to persist. This is ordinarily a weakness that VISReg targets, but our Fourier method may actually benefit from this.

\newpage
\subsection{Auxiliary head in CEM planning}
\label{app:aux-cem}
As discussed earlier, the auxiliary head can be used both indirectly in planning by forcing the encoder to be spatially aware, and directly by physically informing the CEM during planning. We examine the direct method in this section, where the predicted physical state of the auxiliary head can be used to impose physicallly-grounded task-specific constraints on imagined latent trajectories. We choose to examine this setting in the $n$-ball environments, where a ball must be controlled to reach a target while under the constraint of avoiding collision with moving environment balls.

The baseline LeWorldModel planning algorithm uses CEM to minimize the latent distance between an action sequence and a goal.
$$
C_{\text{latent}} = \left\| \hat{z}_H - z_g \right\|_2^2
$$

While the main method explored in our paper improves the physical information present in latent $\hat{z}_H$, thus indirectly physically grounding the CEM, this only places a soft constraint on collision avoidance. As an example, consider two configurations which are visually and spatially similar, but one configuration is near-collision and the other is colliding. Since these configurations are spatially similar, they may be close in latent space, but the impact on CEM success is stepwise. The planner receives a smooth objective from this latent loss, and does not necessarily place a hard constraint on rule-violating trajectories. To address this, we incorporate the auxiliary head prediction during the planning phase. First we reconstruct object locations from the auxiliary objective using a generalized inverse reconstruction map $\Psi_W : y_t^{\text{spec}}\mapsto \{\hat{b}_{t,i}\}_{i=1}^N$:  

$$
\left(\hat{p}_t^{\mathrm{ctrl}},\left\{\hat{p}_{t,j}^{\mathrm{env}}\right\}_{j=1}^{N_{\mathrm{env}}}\right)= \Psi_W A_\psi(\hat{z}_t)
$$
For simplicity in this experiment, we directly train the auxiliary head to predict the positions of the balls rather than the Fourier signature, so $\Psi_W = \operatorname{Id}$. For the Fourier auxiliary head, $\Psi_W$ can either be a simple learned decoder trained using synthetic mapping data, or an algorithmic minimization of plausible Fourier sums. Once the centers of the objects are recovered from the auxiliary head, for balls of radius $r$, we penalize any predicted control ball--environment ball pair whose center separation falls below the collision distance $2r$, 

$$
C_{\mathrm{coll}}=\sum_{t=1}^{H}\sum_{j=1}^{N_{\mathrm{env}}}\operatorname{ReLU}\!\left(2r-\left\|\hat{p}_t^{\mathrm{ctrl}}-\hat{p}_{t,j}^{\mathrm{env}}\right\|_2\right)^2
$$

The auxiliary-informed CEM objective thus becomes 
$$ 
C_{\mathrm{PI\text{-}CEM}}=\left\|\hat{z}_{H}-z_g\right\|_2^2+\lambda_{\mathrm{coll}}C_{\mathrm{coll}}. 
$$
This collision term is solely derived from the model's imagined latent trajectory and auxiliary predictions; no ground-truth ball positions are provided to CEM during planning. Therefore, the auxiliary head provides a predictive and robust way of imposing the known collision constraint on latent-space planning sequences.

\begin{table*}[h]
\centering
\caption{Effect of auxiliary-head-informed CEM in the $n$-ball environments. Lower final distance is better; higher collision-free and success rates are better.}
\label{tab:aux_cem}
\setlength{\tabcolsep}{4pt}
\resizebox{0.98\textwidth}{!}{%
\begin{tabular}{cc cc cc cc}
\toprule
& & \multicolumn{2}{c}{\textbf{Final Dist. $\downarrow$}} & \multicolumn{2}{c}{\textbf{Collision-Free $\uparrow$}} & \multicolumn{2}{c}{\textbf{Success ($r+1$) $\uparrow$}} \\
\cmidrule(lr){3-4}\cmidrule(lr){5-6}\cmidrule(lr){7-8}
\textbf{Env. Balls} & \textbf{Episodes} & \textbf{CEM} & \textbf{Aux-CEM} & \textbf{CEM} & \textbf{Aux-CEM} & \textbf{CEM} & \textbf{Aux-CEM} \\
\midrule
$n=1$ & 20K & \textbf{4.288} & 4.317 & 95.9\% & \textbf{98.8\%} & 90.9\% & \textbf{93.0\%} \\
$n=1$ & 50K & \textbf{3.998} & 4.009 & 96.0\% & \textbf{99.0\%} & 95.1\% & \textbf{97.9\%} \\
\midrule
$n=3$ & 20K & \textbf{5.164} & 5.481 & 85.8\% & \textbf{93.4\%} & 61.8\% & \textbf{64.1\%} \\
$n=3$ & 50K & \textbf{5.074} & 5.347 & 86.0\% & \textbf{92.9\%} & 63.6\% & \textbf{66.2\%} \\
\midrule
$n=5$ & 20K & \textbf{5.550} & 6.278 & 78.4\% & \textbf{88.3\%} & 48.8\% & \textbf{49.1\%} \\
$n=5$ & 50K & \textbf{4.866} & 5.497 & 79.8\% & \textbf{90.6\%} & 63.7\% & \textbf{65.3\%} \\
\bottomrule
\end{tabular}%
}
\end{table*}

We see that the auxiliary-informed CEM improves collision avoidance for all experiments, especially when the number of balls increases. For these higher $n$ environments, the soft collision avoidance in CEM may not be enough to push the selection of imagined trajectories away from a collision sequence. This improvement comes at a slight tradeoff; while terminal distance is virtually similar for $n=1$, for the higher $n$ environments the auxiliary-informed CEM ends up farther from the target location than the baseline CEM on average. Despite this, the collision avoidance provides enough benefit such that overall success rates are higher across all experiments with the auxiliary CEM. We thus show in this experiment that the auxiliary objective can be used both in training and in planning to penalize constraint- or physically-invalid imagined latent trajectories.

\begin{comment}
\Aneesh{State vector as a target (Ablation):
State vector being:
\begin{itemize}
\item Baseline LeWM
\item Exact: basically the state available from the environment: [end-effector x, y, z, end-effector sin(yaw), cos(yaw), gripper opening, gripper contact, cube x, y, z, cube sin(yaw), cos(yaw)]
\item compact7 
\item compact13 adds 3D coordinates
\item Bounding box
\end{itemize}}
\end{comment}
\begin{comment}
\Aneesh{
Compare across both Vit-Tiny and Vit-Small. We show that looking at Vit-Tiny results, hypothesis was that the compact13 captured more information than compact7 and was needed for proper aux head use, but the latent wasn’t big enough to capture the relationship and the model is overdetermined (we are forcing it to learn too much). Therefore, a bigger encoder architecture such as Vit-Small would be able to capture more info and we see that performance increases. This is most notable in OG-Bench cube which doesn't see a performance increase vs Baseline LeWM for ViT-Tiny, which is probably because OG-Bench cube is the most photorealistic out of all of the benchmarks and thus requires a lot out of the encoder which ViT-Tiny is simply too small capacity to handle.}
\end{comment}

\clearpage
\subsection{Target Ablations}
\label{app:target_ablations}

One key ablation is the choice of target for the auxiliary head prediction. These include: 
\begin{itemize}
    \item \textbf{Exact.} The exact environment-specific physical state available from the simulator. 
    \item \textbf{Compact7.} A generalized set of values for each object in the environment: $b=[p_{\text{obj}}, x,y,v_x,v_y,\sin(\theta),\cos(\theta)]$. $p_{\text{obj}}$ is a boolean flag indicating whether the object is currently visible in frame, and any quantities that do not apply are masked when the loss is calculated, often used in bounding box detection models such as DETR~\citep{DBLP:journals/corr/abs-2005-12872}.
    \item \textbf{Compact13.} An expanded version of Compact7 intended for 3D environments, with roll $\phi$, pitch $\psi$, and yaw $\theta$: 
    
    $b=[p_{\text{obj}},x,y,z,vx,vy,vz,\sin(\phi),\cos(\phi),\sin(\psi),\cos(\psi),\sin(\theta),\cos(\theta)]$. 
    \item \textbf{Bounding Boxes.} Our paper's target, regular bounding boxes. 
    \item \textbf{Oriented Bounding Boxes ($\theta$).} Bounding boxes can also be detected in an oriented fashion, with an additional detected angle $\theta$. This changes the target to $b=[x,y,w,h,\sin(\theta),\cos(\theta)]$.
    \item \textbf{Oriented Bounding Boxes ($2\theta$).} Since bounding boxes are rectangular, it is symmetric over a rotation of $\pi$. Therefore, using $2\theta$ can prevent instability from this rotational symmetry. The target is $b=[x,y,w,h,\sin(2\theta),\cos(2\theta)]$.
    \item \textbf{Bounding Boxes + Center of Mass.} Given that bounding boxes are available for the objects, it can be assumed that semantic masks may also be extracted in an unsupervised manner. This can be used to calculate a center of mass as described in Appendix \ref{app:target_creation} that acts as a proxy for the orientation of the object. The target is therefore $b=[x,y,w,h,COM_x,COM_y]$.
    \item \textbf{Oriented Bounding Boxes ($\theta$) + Center of Mass.} We also add the COM to the oriented bounding box target, making it $b=[x,y,w,h,\sin(\theta),\cos(\theta),COM_x,COM_y]$.
    \item \textbf{Oriented Bounding Boxes ($2\theta$) + Center of Mass.} We can also add the COM to the $2\theta$ oriented bounding box target, making it $b=[x,y,w,h,\sin(2\theta),\cos(2\theta),COM_x,COM_y]$.
\end{itemize}

\begin{table}[ht]
\centering
% Increases the vertical spacing between rows slightly
\renewcommand{\arraystretch}{1.2} 
\begin{tabular}{l c c c}
\toprule
\textbf{Target} & \textbf{Push-T} & \textbf{OG-Bench Cube} & \textbf{3-Balls (H=4)} \\
\midrule
Exact                 & 93.6\% & 71.2\% & 77.2\% \\
Compact7              & 89.6\% & 70.4\% & 55.0\% \\
Compact13             & -      & \textbf{74.8\%} & -      \\
BB                    & \textbf{95.2\%} & 70.4\% & \textbf{81.2\%} \\
OBB ($\theta$)        & 94.0\% & 71.2\% & 66.1\% \\
OBB ($2\theta$)       & 94.0\% & 68.4\% & 73.2\% \\
BB + COM              & 93.6\% & 69.6\% & 77.3\% \\
OBB ($\theta$) + COM  & 93.6\% & 67.2\% & 76.3\% \\
OBB ($2\theta$) + COM & 93.2\% & 68.4\% & 77.5\% \\
\bottomrule
\end{tabular}
\caption{Auxiliary target ablation results.}
\label{tab:target_ablation_results}
\end{table}

For the above 2D environments, Push-T and 3-Balls, it is clear that the bounding box target provides the highest success rates. Additional information from oriented bounding boxes or semantic mask centers, or even privileged environment information in the Exact and Compact7 representations do not improve results. For OG-Bench Cube, the 3D representations of the Compact13 target yield the highest success rate, indicating that the encoder is able to capture 3D information when supervised using an appropriate auxiliary head. Even though the 2D bounding box target does change in size and therefore captures a proxy of object depth, it is unable to structure the latent as effectively as explicit 3D state supervision. Regardless, this information may not be available to the model at either training or inference time, and therefore cannot be generalized effectively. 

\newpage
\subsection{Aux Head Architecture Ablations}

The final method uses a shared auxiliary head on both the decoded and predicted latents. We ablate the use of the auxiliary head on the predicted latent, therefore removing the auxiliary loss on the predictor. The encoded loss is simplified accordingly. 

$$\mathcal{L}_{P_{\phi}}=\mathcal{L}_{\text{pred}},
\quad
\mathcal{L}_{E_{\theta}}=\mathcal{L}_{\text{pred}}+\lambda_{A1}\mathcal{L}_{\text{Aux},t}+\lambda_{S}\text{SIGReg}(Z)$$ 

The results indicate that the gradients from the auxiliary head into the predictor enable it to predict better structured latents, improving planning performance over using a single auxiliary head. However, using a single auxiliary head still shows significant improvement over the baseline LeWM success rates.

\begin{table}[ht]
\centering
\renewcommand{\arraystretch}{1.2} 
\begin{tabular}{l c c c}
\toprule
\textbf{Architecture} & \textbf{Push-T} & \textbf{OG-Bench Cube} & \textbf{3-Balls (H=4)} \\
\midrule
Both Aux Heads      & \textbf{95.2\%} & \textbf{70.0\%} & \textbf{81.2\%} \\
Encoded Latent Only & 94.4\% & 69.6\% & 73.6\% \\
LeWM & 91.2\% & 66.4\% & 1.1\% \\
\bottomrule
\end{tabular}
\caption{Performance comparison of auxiliary head architectures.}
\label{tab:aux_head_archis}
\end{table}

\subsection{Object Matching Ablations} 

The fourier-symmetrization is designed to prevent the arbitrary order of auxiliary target predictions of visually similar or identical objects from being incongruent with the ground truth order and incorrectly penalizing the model. We ablate this by removing the fourier-symmetrization altogether, and by implementing the Hungarian algorithm for target-object matching~\citep{https://doi.org/10.1002/nav.3800020109}.

The Hungarian algorithm is used on the set of $N$ objects to find a permutation $\sigma \in \mathfrak{S}_N$ that minimizes cost:
$$\hat{\sigma}=\arg\min_{\sigma \in \mathfrak{S}_N} \sum_{i}^{N} \mathcal{L}_{match}(b_{i,t}, \hat{b}_{\sigma(i,t)})$$
Where $\mathcal{L}_{match}$ is a pair-wise matching cost calculated similarly to $\mathcal{L}_{Aux_{t}}$ on the ground truth and predicted targets. This optimized permutation is then used to calculate the auxiliary loss. 

Two variants of the Hungarian algorithm matching are used. First, a frame-wise matching where the algorithm is used on every frame. Second, a sequence-wise matching where identities are assigned at the beginning of each 4-frame training sequence. 

\begin{table}[ht]
\centering
\renewcommand{\arraystretch}{1.2} 
\begin{tabular}{l c c c}
\toprule
\textbf{Matching} & \textbf{Push-T} & \textbf{OG-Bench Cube} & \textbf{3-Balls (H=4)} \\
\midrule
No Matching        & 94.8\% & 66.0\% & 76.8\% \\
Frame Hungarian    & \textbf{96.4\%} & 65.2\% & 1.7\%  \\
Sequence Hungarian & 94.4\% & 64.0\% & 2.6\%  \\
Fourier            & 95.2\% & \textbf{70.0\%} & \textbf{81.2\%} \\
\bottomrule
\end{tabular}
\caption{Comparison of various object matching strategies.}
\label{tab:matching_strategies}
\end{table}

While the frame-wise Hungarian matching produces an unusually good results for Push-T, it underperforms no-matching for OG-Bench Cube, and 3-Balls collapses despite being an environment where visually similar objects are present. The fourier symmetrization consistently gives better results than no matching at all.

\newpage
\subsection{Depth Ablations}

For 3D environments, depth may also be available using open-source pre-trained video or image depth models, which can then be used in the auxiliary target.  For our experiments, we obtain depth for OG-Bench Cube from the MuJoCo environment itself. For each object $i$, we summarize its camera depth using selected percentiles $z^{(p)}_i$. Using the mean and standard deviation statistics $\mu_p$ and $\sigma_p$ for all the visible objects in the training set, we obtain statistical depth values:
$$\hat{z}^{(p)}_{i}
=
\frac{\log z^{(p)}_{i}-\mu_p}{\sigma_p}$$
We test three training targets:
\begin{itemize}
    \item BB + $\log(\text{median depth})$: $b=[x,y,w,h,\hat{z}^{(50)}]$.
    \item BB + $\log(z_{25}) + \log(z_{75})$: $b=[x,y,w,h,\hat{z}^{(25)},\hat{z}^{(75)}]$.
    \item BB + $\log(z_{05}) + \log(z_{95})$: $b=[x,y,w,h,\hat{z}^{(05)},\hat{z}^{(95)}]$.
\end{itemize}

The results show that adding the 25th and 75th percentile depths marginally improves performance over the bounding box, but does still not provide sufficient 3D geometric information to be as useful as the object centroids in the Compact13 target in ablation \ref{app:target_ablations}.

\begin{table}[ht]
\centering
\renewcommand{\arraystretch}{1.2} 
\begin{tabular}{l c}
\toprule
\textbf{Target} & \textbf{OG-Bench Cube} \\
\midrule
LeWM                               & 66.4\% \\
BB                        & 70.4\% \\
BB + $\log(\text{median depth})$   & 69.6\% \\
BB + $\log(z_{25}) + \log(z_{75})$ & \textbf{70.8\%} \\
BB + $\log(z_{05}) + \log(z_{95})$ & 69.6\% \\
\bottomrule
\end{tabular}
\caption{Depth ablation on OG-Bench Cube.}
\label{tab:depth_ablation}
\end{table}

\subsection{Noisy Bounding Box Ablation}

Detected bounding boxes may be noisy in certain cases. We perturb the ground truth bounding boxes by applying a gaussian jitter to each edge $e$ of up to 1\% of the image width, therefore being 2.24 pixels for the LeWM environments and 0.64 pixels for the n-ball environments.

$$e'=e+\epsilon, \qquad \epsilon\sim\mathcal{N}(0,0.01^2)$$

Results indicate that while noising the bounding boxes decreases performance slightly in some cases, it still improves success rate over the baseline method, and still prevents collapse in the 3-Balls environment, demonstrating the validity of this method with imperfect bounding box annotations.

\begin{table}[ht]
\centering
\renewcommand{\arraystretch}{1.2} 
\begin{tabular}{l c c c}
\toprule
\textbf{Ablation} & \textbf{Push-T} & \textbf{OG-Bench Cube} & \textbf{3-Balls (H=4)} \\
\midrule
Bounding Boxes & \textbf{95.2\%} & \textbf{70.4\%} & \textbf{81.2\%} \\
Noisy Bounding Boxes   & 93.6\% & \textbf{70.4\%} & 75.9\% \\
LeWM        & 91.2\% & 66.4\% & 1.1\%  \\
\bottomrule
\end{tabular}
\caption{Performance comparison between BB baselines and LeWM.}
\label{tab:bb_baseline_lewm}
\end{table}

\newpage
\subsection{Zero-shot Object Detection Pipeline for Bounding Box Data}

To demonstrate that bounding-box supervision does not depend on simulator states or manual annotation, we implement an object discovery, tracking, and bounding-box labeling pipeline using pretrained visual models. Here, \emph{zero-shot} indicates that the pipeline is completely unsupervised, with no environment-specific object
labels, semantic class names, object counts, simulator states, or model
fine-tuning. 

\begin{figure}[h]
    \centering
    \includegraphics[width=0.8\linewidth]{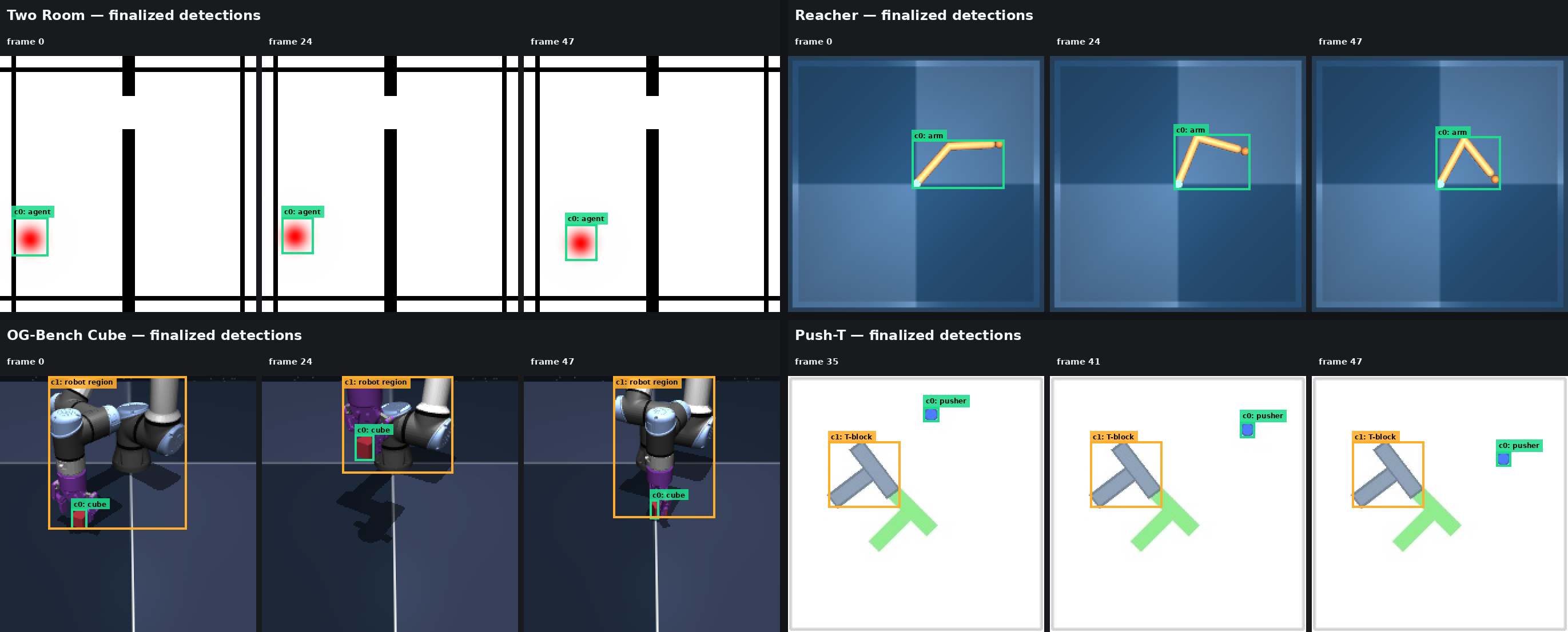}
    \caption{Sample bounding box detections on LeWM environments using the zero-shot object detection pipeline.}
    \label{fig:zeroshotresults}
\end{figure}

For object discovery, we first apply VideoCutLER~\citep{wang2023videocutlersurprisinglysimpleunsupervised} to obtain a collection of object-mask proposals from sparsely sampled video frames. These proposals are then passed through a DINOv2~\citep{oquab2024dinov2learningrobustvisual} encoder to obtain embeddings for each mask. Principal Component Analysis (PCA) and the HDBSCAN~\citep{Malzer_2020} clustering algorithms are used to infer object categories within each environment, producing a local object vocabulary without requiring semantic names or a pre-defined number of objects. 

For consistently tracking the objects and labeling the bounding boxes, we construct an appearance prototype by averaging normalized DINOv2 embeddings from each object over multiple frames. These appearance prototypes are passed into a pretrained Segment Anything 3 (SAM3) video tracker~\citep{carion2026sam3segmentconcepts} that propagates the object masks throughout each sequence of frames. Heuristic filters are applied to combine masks that frequently overlap and remove masks deemed relevant by a variety of pre-determined, environment-agnostic factors. If a retained category is missing from a frame, the SAM3 video tracker is re-initialized with the object's DINOv2 embedding to re-acquire its position in subsequent frames.

Finally, the masks are converted into the axis-aligned bounding box representation by finding the bounds of the tracked semantic mask for each frame. These are stored along with the inferred object category and instance identifier for each clip. These masks can also be converted deterministically into oriented bounding boxes and can be used to calculate centers of mass using similar geometric post-processing techniques. 

Sample bounding box detections are shown in Figure \ref{fig:zeroshotresults} using this zero-shot object detection pipeline, which closely resemble the simulator ground truth labels shown in Appendix \ref{app:target_creation}. Though the principal experiments use bounding box labels derived from the simulator, this is used as a demonstration of feasibility for a generalized, unsupervised application of this auxiliary head method without requiring manual frame-level annotation or a simulator to provide the bounding box labels.

% \subsection{Slot Attention Auxiliary Head}

\newpage
\subsection{Linear and MLP Probing}
Following the LeWM evaluation protocol, and inspired by \cite{joseph2026interpreting} probing, we train additional linear and MLP probes to explicitly quantify how well physical quantities can be extracted from latent embeddings. Better performance on linear probes indicate a better linear representation of the physical quantity within the latent space, while higher performance on an MLP probe indicates a more entangled representation. 

\textbf{Implementation Details}. Probes are trained using 50000 latents from each environment, and tested on 5000 latents. The linear probe is trained using ridge regression with $\lambda=10^{-6}$. The MLP probe uses a simple 2-layer MLP, projecting the latent to a 256-D feature before predicting the output state. It is trained for 100 epochs with a batch size of 1024. They are initialized with seeds 3072, 3073, and 3074, and their outputs are averaged for the final results. Probes are trained for each individual state, which are determined by the environments themselves. We report the mean squared error (MSE), and the Pearson correlation coefficient
between the ground-truth and predicted states.

\textbf{Results}. Table \ref{tab:combined_probe_results} shows improvements in all environments for for most quantities, demonstrating the effectiveness of the auxiliary target in better organizing the latent space with respect to key physical properties. Push-T and the 6-ball environment benefit the most, while 1-ball and 3-ball environments demonstrate a decrease in the prediction performance of the environment balls' positions, which is in line with the reduced spearman correlation in the main results. Two-Room shows poorer linear probe performance, but improved MLP probe performance, indicating the auxiliary target results in an entangled, but improved encoding of positional information in the latent space. 

\begin{table*}[t]
\centering
\caption{Linear and MLP probe results for the final models.
Ours uses Fourier-supervised bounding-box prediction with the predictor-side
auxiliary head frozen.
The shaded rows are the primary metrics.
Bold indicates lower MSE or higher Pearson correlation $r$.}
\label{tab:combined_probe_results}

\setlength{\tabcolsep}{4pt}
\renewcommand{\arraystretch}{1.12}

\resizebox{\textwidth}{!}{%
\begin{tabular}{llcccccccc}
\toprule
\textbf{Environment}
& \textbf{Target}
& \multicolumn{2}{c}{\textbf{Linear MSE $\downarrow$}}
& \multicolumn{2}{c}{\textbf{Linear $r$ $\uparrow$}}
& \multicolumn{2}{c}{\textbf{MLP MSE $\downarrow$}}
& \multicolumn{2}{c}{\textbf{MLP $r$ $\uparrow$}} \\
\cmidrule(lr){3-4}
\cmidrule(lr){5-6}
\cmidrule(lr){7-8}
\cmidrule(lr){9-10}
&
& \textbf{Ours} & \textbf{LeWM}
& \textbf{Ours} & \textbf{LeWM}
& \textbf{Ours} & \textbf{LeWM}
& \textbf{Ours} & \textbf{LeWM} \\
\midrule

% Push-T
\rowcolor{gray!15}
\cellcolor{white}\multirow{4}{*}{\textbf{Push-T}}
& \textbf{Full state (primary)}
& \textbf{0.02998} & 0.05672
& \textbf{0.98484} & 0.97108
& 0.00576 & \textbf{0.00547}
& 0.99714 & \textbf{0.99729} \\
& Pusher XY
& \textbf{0.02203} & 0.05629
& \textbf{0.98900} & 0.97160
& \textbf{0.00981} & 0.01014
& \textbf{0.99515} & 0.99500 \\
& T-block XY
& \textbf{0.01568} & 0.02598
& \textbf{0.99224} & 0.98711
& 0.00123 & \textbf{0.00108}
& 0.99941 & \textbf{0.99948} \\
& T-block yaw
& \textbf{0.05221} & 0.08790
& \textbf{0.97327} & 0.95453
& 0.00624 & \textbf{0.00519}
& 0.99686 & \textbf{0.99738} \\

\addlinespace[0.65em]

% Two Room
\rowcolor{gray!15}
\cellcolor{white}\textbf{Two Room}
& \textbf{Full state (primary)}
& \textbf{0.00259} & 0.00379
& \textbf{0.99870} & 0.99811
& \textbf{0.00012} & 0.00018
& \textbf{0.99994} & 0.99992 \\

\addlinespace[0.65em]

% Reacher
\rowcolor{gray!15}
\cellcolor{white}\multirow{3}{*}{\textbf{Reacher}}
& \textbf{Full state (primary)}
& \textbf{0.22869} & 0.22958
& \textbf{0.87226} & 0.87172
& \textbf{0.22789} & 0.22789
& 0.87268 & \textbf{0.87269} \\
& Joint angles
& \textbf{0.45409} & 0.45531
& \textbf{0.74620} & 0.74541
& \textbf{0.45570} & 0.45571
& 0.74540 & \textbf{0.74540} \\
& Finger location
& \textbf{0.00329} & 0.00386
& \textbf{0.99833} & 0.99804
& 0.00007 & \textbf{0.00006}
& 0.99997 & \textbf{0.99997} \\

\addlinespace[0.65em]

% OG-Bench Cube
\rowcolor{gray!15}
\cellcolor{white}\multirow{7}{*}{\textbf{OG-Bench Cube}}
& \textbf{Full state (primary)}
& \textbf{0.36245} & 0.36666
& \textbf{0.66975} & 0.66651
& 0.39748 & \textbf{0.39610}
& \textbf{0.66696} & 0.66414 \\
& Effector location
& \textbf{0.00946} & 0.01668
& \textbf{0.99521} & 0.99154
& \textbf{0.00203} & 0.00255
& \textbf{0.99898} & 0.99871 \\
& Effector yaw
& \textbf{0.99748} & 0.99850
& \textbf{0.10474} & 0.09879
& 1.14190 & \textbf{1.13966}
& \textbf{0.05170} & 0.04293 \\
& Gripper opening
& \textbf{0.10291} & 0.11178
& \textbf{0.94705} & 0.94234
& \textbf{0.03405} & 0.03922
& \textbf{0.98286} & 0.98033 \\
& Gripper contact
& \textbf{0.18313} & 0.18816
& \textbf{0.90385} & 0.90105
& \textbf{0.10303} & 0.10856
& \textbf{0.94722} & 0.94440 \\
& Cube location
& \textbf{0.00715} & 0.01143
& \textbf{0.99639} & 0.99422
& \textbf{0.00146} & 0.00183
& \textbf{0.99927} & 0.99908 \\
& Cube yaw
& 1.00929 & \textbf{1.00929}
& \textbf{0.00088} & -0.00009
& 1.16924 & \textbf{1.15646}
& \textbf{-0.01238} & -0.01713 \\

\addlinespace[0.65em]

% n = 1
\rowcolor{gray!15}
\cellcolor{white}\multirow{3}{*}{\textbf{$n=1$ balls}}
& \textbf{Full positions (primary)}
& \textbf{0.00859} & 0.49398
& \textbf{0.99572} & 0.71278
& \textbf{0.00053} & 0.50527
& \textbf{0.99974} & 0.70494 \\
& Control ball
& \textbf{0.00637} & 0.98284
& \textbf{0.99684} & 0.15859
& \textbf{0.00042} & 1.00990
& \textbf{0.99979} & 0.05758 \\
& Environment balls
& 0.01080 & \textbf{0.00512}
& 0.99459 & \textbf{0.99744}
& 0.00043 & \textbf{0.00020}
& 0.99979 & \textbf{0.99990} \\

\addlinespace[0.65em]

% n = 3
\rowcolor{gray!15}
\cellcolor{white}\multirow{3}{*}{\textbf{$n=3$ balls}}
& \textbf{Full positions (primary)}
& \textbf{0.19330} & 0.36222
& \textbf{0.89821} & 0.79862
& \textbf{0.10821} & 0.33956
& \textbf{0.94443} & 0.81328 \\
& Control ball
& \textbf{0.00726} & 0.96038
& \textbf{0.99643} & 0.23830
& \textbf{0.00040} & 1.10246
& \textbf{0.99981} & 0.11921 \\
& Environment balls
& 0.25531 & \textbf{0.16283}
& 0.86209 & \textbf{0.91443}
& 0.14405 & \textbf{0.09313}
& 0.92483 & \textbf{0.95201} \\

\addlinespace[0.65em]

% n = 6
\rowcolor{gray!15}
\cellcolor{white}\multirow{3}{*}{\textbf{$n=6$ balls}}
& \textbf{Full positions (primary)}
& \textbf{0.47547} & 0.95494
& \textbf{0.72685} & 0.23115
& \textbf{0.48770} & 1.05168
& \textbf{0.71948} & 0.14460 \\
& Control ball
& \textbf{0.00527} & 0.83346
& \textbf{0.99740} & 0.42038
& \textbf{0.00039} & 0.96106
& \textbf{0.99981} & 0.33427 \\
& Environment balls
& \textbf{0.55384} & 0.97519
& \textbf{0.67093} & 0.18149
& \textbf{0.56514} & 1.08073
& \textbf{0.66404} & 0.09657 \\

\bottomrule
\end{tabular}%
}
\end{table*}

\end{document}